\documentclass{article}
\usepackage{arxiv}

\usepackage{times}
\usepackage[numbers]{natbib}
\usepackage{multicol}
\usepackage[bookmarks=true]{hyperref}
\usepackage{microtype}
\usepackage{bm}
\usepackage{MnSymbol}
\usepackage{mathdots}
\usepackage{amsthm}
\usepackage{amsfonts}
\usepackage{empheq}
\usepackage{algorithm}
\usepackage{algpseudocode}
\usepackage{siunitx}
\usepackage{tikz}
\usepackage{makecell}
\usepackage[caption=false,font=footnotesize]{subfig}
\usepackage{multirow}
\usepackage{graphicx}
\usepackage{booktabs}

\theoremstyle{plain}

\theoremstyle{definition}

\theoremstyle{remark}

\newcommand{\R}{\mathbb{R}}

\providecommand{\dt}{\mathrm{dt}}

\newcommand{\inv}[1]{{#1}^{-1}}

\newcommand{\grpG}{\mathbf{G}}
\newcommand{\gothg}{\mathfrak{g}}

\newcommand{\twoel}[2]{\left(#1,\; #2\right)}
\newcommand{\threeel}[3]{\left(#1,\; #2,\; #3\right)}
\newcommand{\fourel}[4]{\left(#1,\; #2,\; #3,\; #4\right)}

\newcommand{\Adsym}[2]{\mathrm{Ad}_{#1}\!\left[#2\right]}

\newcommand{\AdMsym}[1]{\mathbf{Ad}^\vee_{#1}}

\newcommand{\SO}[1]{\mathbf{SO}(#1)}
\newcommand{\so}[1]{\mathfrak{so}(#1)}

\newcommand{\G}[1]{\mathbf{Gal}(#1)}
\newcommand{\g}[1]{\mathfrak{gal}(#1)}

\newcommand{\Jl}[1]{\mathbf{J}_{\mathrm{L}}(#1)}

\newcommand{\diff}[1]{\mathrm{d}{#1}}

\newcommand{\Vector}[3]{\prescript{#1}{}{\bm{#2}}_{#3}}

\newcommand{\hatVector}[3]{\prescript{#1}{}{\hat{\bm{#2}}}_{#3}}

\newcommand{\barVector}[3]{\prescript{#1}{}{\bar{\bm{#2}}}_{#3}}

\newcommand{\ringVector}[3]{\prescript{#1}{}{\mathring{\bm{#2}}}_{#3}}

\newcommand{\hatbias}[2]{\hatVector{#1}{b}{#2}}

\newcommand{\tauInp}[2]{\Vector{#1}{\tau}{#2}}
\newcommand{\omegaInp}[2]{\Vector{#1}{\omega}{#2}}
\newcommand{\aInp}[2]{\Vector{#1}{a}{#2}}

\newcommand{\eyen}[1]{\mathbf{I}_{#1}}

\newcommand{\zeronm}[2]{\mathbf{0}_{#1 \times #2}}

\newcommand{\Rn}[1]{\R^{#1}}

\newcommand{\Rnm}[2]{\R^{#1 \times #2}}

\newcommand{\point}{\;\cdot\;}

\newcommand{\Lift}[1]{\Lambda_{\scalebox{0.5}{$#1$}}}

\newcommand{\gn}{\Vector{}{g}{\scalebox{0.5}{\textnormal{N}}}}

\newcommand{\wn}{\Vector{}{w}{\scalebox{0.5}{\textnormal{N}}}}

\newcommand{\ringwn}{\ringVector{}{w}{\scalebox{0.5}{\textnormal{N}}}}

\newcommand{\Matrix}[3]{{^{#1}\mathbf{#2}_{#3}}}

\newcommand{\dotMatrix}[3]{{^{#1}\dot{\mathbf{#2}}_{#3}}}

\newcommand{\hatMatrix}[3]{{^{#1}\hat{\mathbf{#2}}_{#3}}}

\newcommand{\figref}[1]{Fig.~\ref{fig:#1}}
\newcommand{\secref}[1]{Sec.~\ref{sec:#1}}

\newcommand{\tabref}[1]{Tab.~\ref{tab:#1}}
\newcommand{\algoref}[1]{Alg.~\ref{alg:#1}}

\newcommand{\papercitation}{
G. Delama, M. Scheiber, Y. Ge, T. Hamel, S. Weiss and R. Mahony, ``Galilean State Estimation for Inertial Navigation
Systems with Unknown Time Delay'', in Proceedings of \emph{Robotics: Science and Systems (RSS)}, 2026.
}

\newcommand{\publicationdetails}
{
\copyrightNoticeRSSAccepted{2026} 
\papercitation 
}
\newcommand{\publicationversion}
{Author accepted version}

\begin{document}

\title{Galilean State Estimation for Inertial Navigation Systems with Unknown Time Delay}
\headertitle{Galilean State Estimation for Inertial Navigation Systems with Unknown Time Delay}

\author{
\href{https://orcid.org/0009-0007-3918-9196}{\includegraphics[scale=0.06]{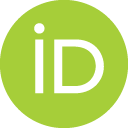}\hspace{1mm}
Giulio Delama}\\
	\href{https://www.aau.at/en/smart-systems-technologies/control-of-networked-systems/}{Control of Networked Systems Group} \\
	University of Klagenfurt \\
    9020 Klagenfurt, Austria \\
	\texttt{\href{mailto:giulio.delama@ieee.org}{giulio.delama@ieee.org}} \\
\And \href{https://orcid.org/0000-0002-3415-7378}{\includegraphics[scale=0.06]{orcid.png}\hspace{1mm}
 Martin Scheiber}\\
	\href{https://www.aau.at/en/smart-systems-technologies/control-of-networked-systems/}{Control of Networked Systems Group} \\
	University of Klagenfurt \\
    9020 Klagenfurt, Austria \\
	\texttt{\href{mailto:martin.scheiber@ieee.org}{martin.scheiber@ieee.org}} \\
\And \href{https://orcid.org/0000-0001-7969-7039}{\includegraphics[scale=0.06]{orcid.png}\hspace{1mm}
 Yixiao Ge}\\
	Systems Theory and Robotics Group \\
	Australian National University \\
    ACT, 2601, Australia \\
	\texttt{\href{mailto:yixiao.ge@anu.edu.au}{yixiao.ge@anu.edu.au}} \\
\And \href{https://orcid.org/0000-0002-7779-1264}{\includegraphics[scale=0.06]{orcid.png}\hspace{1mm}
 Tarek Hamel}\\
	I3S, CNRS, Université Côte d’Azur \\
	and Institut Universitaire de France \\
    06900 Sophia Antipolis, France \\
	\texttt{\href{mailto:thamel@i3s.unice.fr}{thamel@i3s.unice.fr}} \\
\And \href{https://orcid.org/0000-0001-6906-5409}{\includegraphics[scale=0.06]{orcid.png}\hspace{1mm}
Stephan Weiss}\\
	\href{https://www.aau.at/en/smart-systems-technologies/control-of-networked-systems/}{Control of Networked Systems Group} \\
	University of Klagenfurt \\
    9020 Klagenfurt, Austria \\
	\texttt{\href{mailto:stephan.weiss@ieee.org}{stephan.weiss@ieee.org}} \\
\And \href{https://orcid.org/0000-0002-7803-2868}{\includegraphics[scale=0.06]{orcid.png}\hspace{1mm}
 Robert Mahony}\\
	Systems Theory and Robotics Group \\
	Australian National University \\
    ACT, 2601, Australia \\
	\texttt{\href{mailto:robert.mahony@anu.edu.au}{robert.mahony@anu.edu.au}} \\
}
\maketitle

\begin{abstract}
Many Inertial Navigation Systems (INS) use Global Navigation Satellite System (GNSS) position as the primary measurement to drive filter performance and bound error growth.
However, commercial-grade GNSS receivers introduce unknown measurement delays ranging from 50\,ms to 300\,ms depending on sensor quality and operating mode.
Such time delays can significantly degrade INS performance unless they are explicitly compensated for.
Existing algorithms commonly estimate this delay offline, run the filter concurrently with GNSS measurements using buffered Inertial Measurement Unit (IMU) data, and predict the current state by forward-integrating buffered inertial measurements via IMU preintegration.
The state-of-the-art online method is an Extended Kalman Filter (EKF) that explicitly models the time delay as a state parameter, which defines the preintegration duration.
This paper introduces a novel geometric framework for modeling time-delayed INS, in which Galilean symmetry is leveraged to provide a joint representation of space and time for consistent state estimation.
An Equivariant Filter (EqF) is derived for the coupled estimation of navigation states and time delay.
Validation is performed on two fixed-wing Uncrewed Aerial Vehicles (UAV) with GNSS time lags of 90\,ms and 120\,ms.
The test flights last two to three minutes.
Simulations further investigate delays up to 500\,ms and provide a statistical comparison against the state-of-the-art EKF.
Results show that the EqF preserves accuracy and consistency, while the EKF lacks consistency and its performance degrades significantly with increasing measurement delays.
\end{abstract}

\keywords{
Galilean Symmetry, Time Delay, Inertial Navigation System, State Estimation.
}

\section{Introduction and Related Work}
\label{sec:introrelated}
Inertial Navigation Systems (INS) estimate the position, velocity, and orientation of a vehicle by fusing high-rate accelerometer and gyroscope measurements from an Inertial Measurement Unit (IMU) with data from other sensors, such as Global Navigation Satellite Systems (GNSS), magnetometers, barometers, visual- or range-based motion sensors.
These systems are widely used in robotic, aerospace, marine, and automotive applications.
However, delays in sensor measurements due to sensor processing, communication latency, or asynchronous sampling introduce measurement uncertainty that compromises INS accuracy. 
Measurements are subject to both \emph{lag}, the systematic delay between the actual measurement and the moment it becomes available for processing, and \emph{jitter}, the smaller, variable fluctuations in arrival time caused by factors such as bus congestion or communication load.
For example, position measurements from commercial-grade GNSS typically have lags in the range of 50\,ms\,-\,300\,ms associated with the coherent integration time of the receiver, and jitter in the order of 20\,ms\,-\,30\,ms\, associated with thread priority in the avionics system. 
Although algorithms exist to estimate and compensate for jitter, accurately determining the time lag is more challenging.
Even small delays can introduce biases and inconsistencies in the estimator, emphasizing the need for INS algorithms that account for these timing offsets.

Several delay-aware state estimation methods have been developed.
Classical optimization‑based techniques estimate both navigation states and sensor delays by iteratively minimizing measurement residuals over batch data. 
Examples include a general multi‑sensor temporal calibration by~\citet{Kelly2014ASensors}, and camera–IMU systems from~\citet{Mingyang2014OnlineAlgorithms},~\citet{Kelly2014DeterminingMeasurements}, and~\citet{Song2025unleashingpowerdiscretetimestate}.
On the other hand,~\citet{Larsen1998IncorporationFilter} and~\citet{Nilsson2010JointSystems} introduced Extended Kalman Filters (EKF) that incorporate delayed measurements by performing extrapolation and interpolation to manage asynchronous updates.
\citet{Fatehi2017KalmanDelay} studied EKF approaches for multi‑rate fusion with irregular sampling and variable measurement delays, highlighting the importance of algorithms that can fuse slow-rate, delayed data with fast measurements in a statistically principled way.
\citet{Frei2023ARendezvous} developed a robust navigation filter for fusing delayed, asynchronous measurements from multiple sensors in spacecraft rendezvous, modifying the EKF update to fuse delayed data effectively in practice.
\citet{Khosravian2016StateMeasurements} proposed an observer–predictor structure for invariant systems with delayed outputs, using geometric properties to combine delayed measurements with state predictions.
More recently,~\citet{vanGoor2024ConstructiveMeasurements} developed a nonlinear INS observer that compensates for temporal offsets in position measurements using recursively computed delay matrices, achieving almost-global asymptotic and locally exponential stability.
These methods focus on processing asynchronous and delayed data within a given filter, rather than explicitly estimating unknown delays online.
\begin{figure}[b]
    \centering

    \resizebox{0.6\columnwidth}{!}{%
    \begin{minipage}{\columnwidth}

        \centering

        \includegraphics[width=\columnwidth]{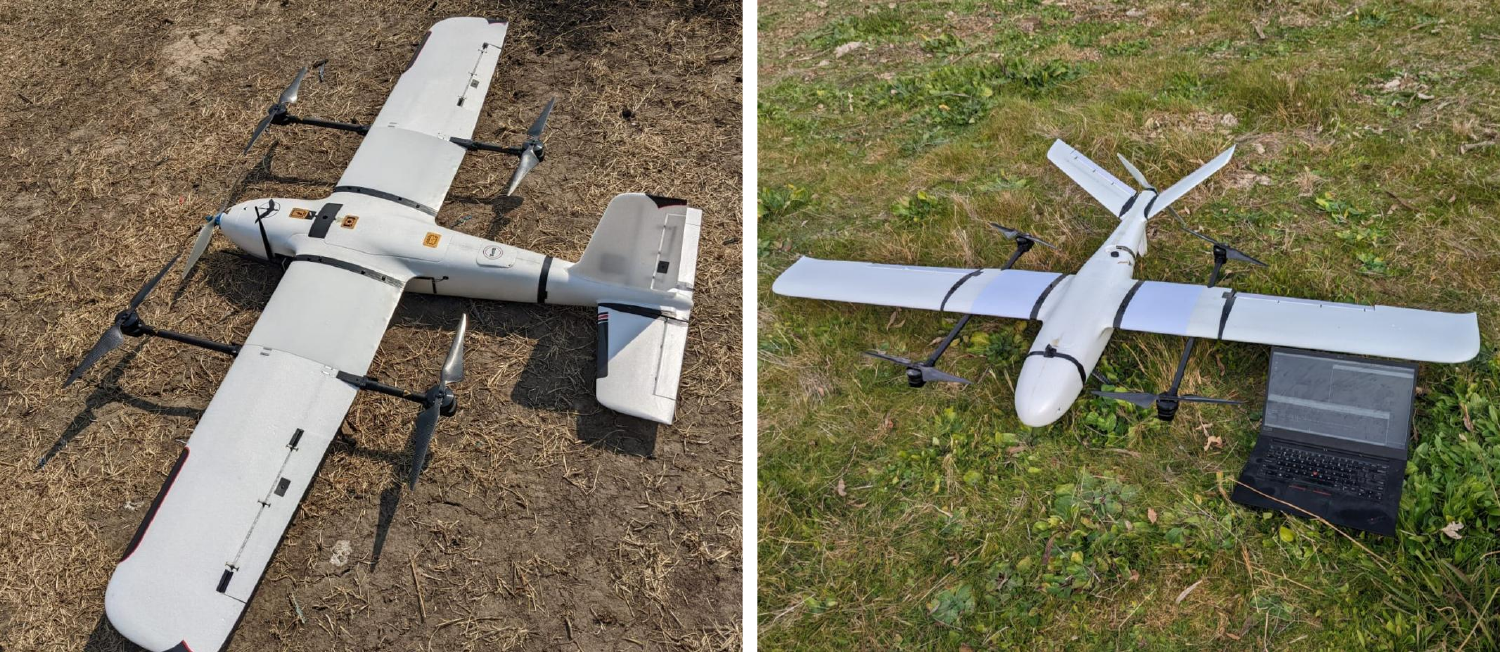}\\[-2mm]

        \subfloat[\texttt{striver}]{%
            \begin{minipage}[b]{0.33\columnwidth}
                \includegraphics[width=\textwidth]{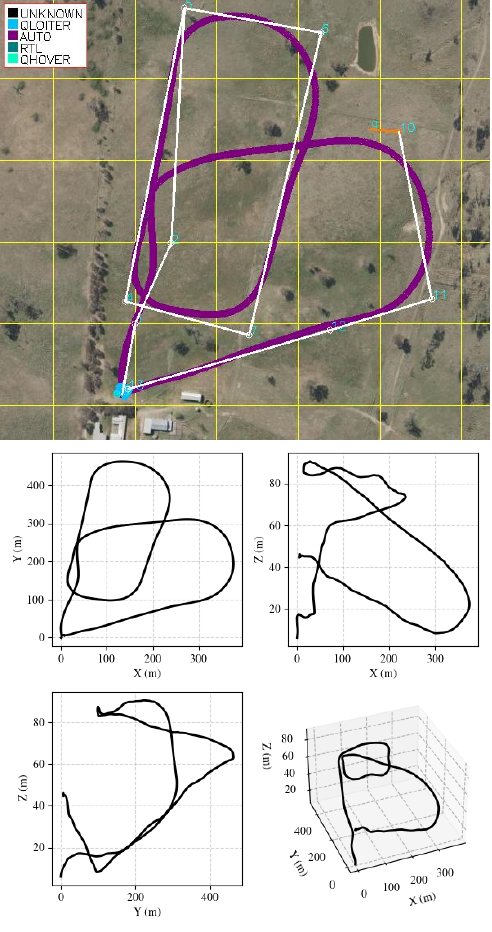}
            \end{minipage}%
        }
        \hfill
        \subfloat[\texttt{hero-1}]{%
            \begin{minipage}[b]{0.33\columnwidth}
                \includegraphics[width=\textwidth]{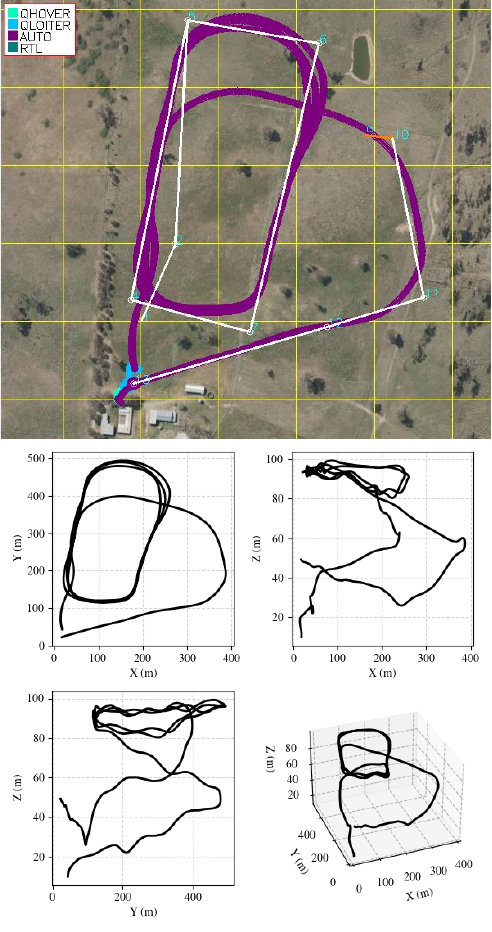}
            \end{minipage}%
        }
        \hfill
        \subfloat[\texttt{hero-2}]{%
            \begin{minipage}[b]{0.33\columnwidth}
                \includegraphics[width=\textwidth]{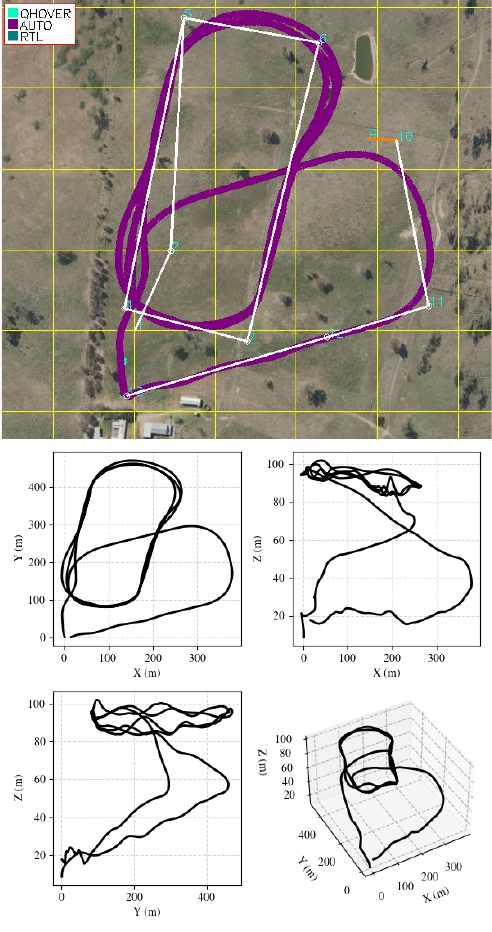}
            \end{minipage}%
        }

    \end{minipage}%
    }

    \caption{Overview of the real-world flight experiments. The top row shows the two fixed-wing UAVs used for data collection, the Makeflyeasy \textit{Striver} (left) and \textit{Hero} (right). The remaining panels show the trajectories of the three dataset sequences (a)~\texttt{striver}, (b)~\texttt{hero-1}, and (c)~\texttt{hero-2}, visualized in map view (middle row) and as 3D trajectory reconstructions (bottom rows).}
    \label{fig:4_overview}
\end{figure}

\citet{Kim2025EKF-BasedCalibration},~\citet{Guo2023EnhancedIntegration}, and~\citet{Mina2025RemarksFiltering} proposed EKF‑based methods to estimate measurement time delays online, jointly updating navigation states and the time-delay state during recursive propagation.
However, while these approaches can improve accuracy in specific applications,~\citet{Kelly2021ASystems} show that including online time-delay estimation in recursive filters introduces fundamental structural issues that lead to bias and inconsistency, which cannot be fully mitigated by tuning filter noise variances.
To the authors' knowledge, there is no existing filter-based online method that does not suffer from the inherent inconsistencies and instabilities identified by~\citet{Kelly2021ASystems}.

Recent research has shown the advantages of exploiting equivariant symmetries in state estimation.
The Equivariant Filter (EqF) by~\citet{vanGoor2022EquivariantEqF} has been successfully applied to a wide range of navigation problems.
Relevant to this work,~\citet{Ge2022EquivariantSystems} extended the EqF to general discrete-time systems, while~\citet{Fornasier2022OvercomingCalibration, Fornasier2022EquivariantBiases, Fornasier2023MSCEqF:Navigation, Fornasier2024AnSystem, Fornasier2025EquivariantSystems} leveraged it for biased-INS, establishing equivariant symmetries that couple navigation states and IMU biases.
\citet{Scheiber2023RevisitingApproach} applied it to multi-GNSS navigation.
By preserving the intrinsic geometry of the state space, the EqF outperforms the classical EKF, providing a larger basin of attraction with faster convergence, improved accuracy, and consistency.

In this paper, we leverage \emph{Galilean symmetry} to develop a geometric framework that enables consistent online estimation of navigation states and unknown time delays in INS.
Galilean symmetry captures the intrinsic spatio-temporal symmetry of the inertial navigation problem and naturally provides a formulation that couples temporal offset with position and velocity states. 
Recently,~\citet{Kelly2023AllSGal3} formalized the \emph{Galilean matrix Lie group} in a technical report, while~\citet{Delama2025EquivariantApproach} leveraged Galilean symmetry to model discrete-time INS and develop a novel \emph{IMU preintegration} method improving consistency and reducing linearization errors.
\citet{Mahony2025GalileanRobotics} provided a robotics‑oriented exposition applying the Galilean group to inertial navigation, kinematics, and sensor fusion under \emph{temporal uncertainty}.
The Galilean formulation allows us to formalize the INS state estimation problem for a general scenario with delayed measurements that have both lag and jitter.
We use this framework to derive an EqF that jointly estimates navigation states and time delays for GNSS-IMU navigation.
Real-world flight experiments with two fixed-wing Uncrewed Aerial Vehicles (UAV) are carried out to compare the performance of the proposed EqF against EKFs with and without offline delay compensation and an EKF with online delay estimation.
For GNSS delays of 90\,ms and 120\,ms, the EqF accurately estimates both navigation states and delays online, outperforming the EKF with delay compensation (where the delay estimate is computed using offline batch optimization).
To provide further statistical analysis, we conduct Monte Carlo simulations with delays up to 500\,ms, showing that the EqF maintains accurate and consistent estimation, whereas the state-of-the-art EKF with online time-delay estimation lacks consistency and exhibits unstable behavior.

The contributions of this paper can be outlined as follows:
\begin{itemize}
    \item A geometric framework based on Galilean symmetry for modeling inertial motion under gravity for multiple rigid bodies with relative temporal offsets, allowing spatio-temporal state estimation and enabling a principled representation of INS with delayed measurements (\secref{problem}).
    \item Application of this framework to a minimal IMU-GNSS setup with delayed position measurements and derivation of an EqF for online joint estimation of navigation states and unknown GNSS time delays (\secref{filter}).
    \item Experimental evaluation of the proposed EqF with real-world UAV flights, and with Monte Carlo simulations providing a statistical comparison against a state-of-the-art EKF and showing improved consistency and accuracy across a range of GNSS time delays (\secref{results}).
\end{itemize}
\section{Galilean Spatio-Temporal State Estimation}
\label{sec:problem}
This section presents a geometric framework for modeling inertial motion with temporal offsets using \emph{Galilean symmetry}. 
Notation and basic concepts for representing spatio-temporal transformations between inertial frames are given in~\secref{notation}, and delayed inertial frames for rigid bodies under gravity are modeled in~\secref{delayed_frames}, distinguishing \emph{intra-body} and \emph{cross-body} transformations. 
Their \emph{continuous-time kinematics} are derived in~\secref{kinematics}, and~\secref{state_estimation} outlines how the framework can be applied to state estimation with delayed measurements.

\subsection{Notation and Mathematical Preliminaries}
\label{sec:notation}
Vectors are denoted by bold lowercase letters $\Vector{}{x}{}$, matrices by bold capital letters $\Matrix{}{X}{}$, and group elements by regular letters $X$ or $x$.
The $n$-dimensional identity matrix is ${\eyen{n} \in \Rnm{n}{n}}$, and the ${n \times m}$ zero matrix is ${\zeronm{n}{m} \in \Rnm{n}{m}}$.
Coordinate frames are denoted by capital letters inside curly brackets $\{X\}$.
In this study, frames of reference at different times are considered.
A subscript $t$ on a frame $\{X_t\}$ specifies the associated time.
Calligraphic symbols $\mathcal{M}$ denote manifolds, e.g., $\mathcal{SO}(3)$, while bold symbols $\mathbf{G}$ denote Lie groups, e.g., $\SO{3}$.

Inertial frames are reference frames subject to Newton's laws of motion.
They are defined by spatial coordinates (orientation, position, and velocity) and a temporal coordinate (time).
The set of all inertial frames is related by \emph{Galilean symmetry}, capturing the invariance of Newtonian mechanics under frame transformations.
A \emph{Galilean frame} (or \emph{transformation}) is a map that expresses the coordinates of one inertial frame in terms of another.
Specifically, for two inertial frames $\{A\}$ and $\{B\}$, the Galilean frame ${\Matrix{A}{F}{B} \in \mathcal{G}al(3)}$ expressed in homogeneous coordinates is represented as
\begin{align}
    \Matrix{A}{F}{B} =
    \begin{bmatrix}
        \Matrix{A}{R}{B} & \Vector{A}{v}{B} & \Vector{A}{p}{B} \\
        \zeronm{1}{3} & 1 & ^{A}t_B \\
        \zeronm{1}{3} & 0 & 1
    \end{bmatrix} \in \Rnm{5}{5},
\end{align}
where ${\Matrix{A}{R}{B} \in \mathcal{SO}(3)}$ is a rotation matrix encoding the relative orientation, ${\Vector{A}{v}{B} \in \Rn{3}}$ the relative \emph{inertial} velocity, ${\Vector{A}{p}{B} \in \Rn{3}}$ the relative position, and ${^{A}t_B \in \R}$ the time offset between the frames.
These homogeneous matrices define the \emph{Galilean matrix Lie group} $\G{3}$ that provides a principled framework for representing coupled spatial and temporal relationships between inertial frames.
An element of the Galilean group ${X \in \G{3}}$ and of its Lie algebra ${U \in \g{3}}$ are given by
\begin{align}
    &X =
    \begin{bmatrix}
        A & a & b \\
        \zeronm{1}{3} & 1 & c \\
        \zeronm{1}{3} & 0 & 1
    \end{bmatrix} \in \Rnm{5}{5}, 
    &&U =
    \begin{bmatrix}
        \Omega & v & r \\
        \zeronm{1}{3} & 0 & \tau \\
        \zeronm{1}{3} & 0 & 0
    \end{bmatrix} \in \Rnm{5}{5} ,
\end{align}
where ${A \in \SO{3}}$, ${\Omega \in \so{3}}$, ${a, b, v, r \in \Rn{3}}$, and ${c, \tau \in \R}$.

The Galilean group identity element is ${I \equiv \eyen{5}}$, and the inverse is ${X^{-1} = \fourel{A^{-1}}{-A^{-1}a}{-A^{-1}(b-ca)}{-c}}$.
The \emph{wedge operator} ${(\cdot)^\wedge : \Rn{n} \to \gothg}$ maps a vector ${\Vector{}{x}{} \in \Rn{n}}$ to the Lie algebra element ${\Vector{}{x}{}^\wedge \in \gothg}$, with inverse ${(\cdot)^\vee : \gothg \to \Rn{n}}$, where ${n = \dim(\gothg)}$.
The action of Galilean symmetry on a Galilean frame ${\Matrix{}{F}{} = \fourel{\Matrix{}{R}{}}{\Vector{}{v}{}}{\Vector{}{p}{}}{t}}$ is given by the matrix multiplication
\begin{align}
    X \Matrix{}{F}{} =
    \begin{bmatrix}
        A \Matrix{}{R}{} & A \Vector{}{v}{} + a & A \Vector{}{p}{} + a \, t + b \\
        \zeronm{1}{3} & 1 & t + c \\
        \zeronm{1}{3} & 0 & 1
    \end{bmatrix} .
\end{align}

\subsection{Galilean Symmetry for Modeling INS with Delay}
\label{sec:delayed_frames}
Accurate modeling of inertial motion requires careful choice of reference frames.
In many robotics applications, the natural reference frame is generally fixed to the Earth’s surface, a non-inertial frame due to gravity and the Earth's rotation.
While the Earth's rotation can often be neglected in many robotics problems, gravity introduces a continuous acceleration $\Vector{}{g}{} \in \Rn{3}$ that must be modeled.
In this context, Galilean symmetry provides a principled and rigorous way to understand the effects of gravity in modeling relative non-inertial motion \cite{Mahony2025GalileanRobotics}.
\begin{figure}[b]
    \centering
    \includegraphics[width=0.6\linewidth]{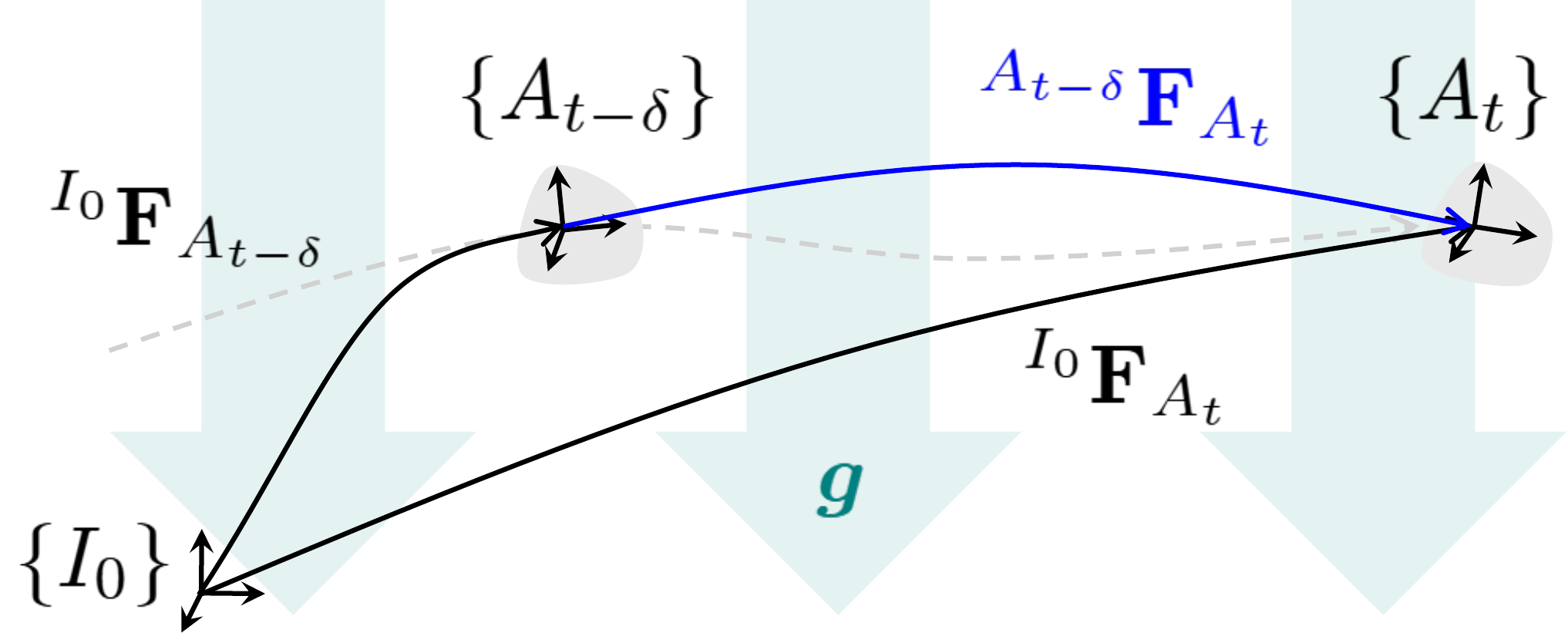}\\
    \caption{Galilean transformations for a rigid body $A$ in a gravity field. $\{I_0\}$ is an inertial frame of reference, defined as an arbitrary free-falling Galilean frame at zero time. The transformations $\Matrix{I_0}{F}{A_{t-\delta}}$ and $\Matrix{I_0}{F}{A_t}$ map \emph{spatial} and \emph{temporal} coordinates from the body frame at times $t-\delta$ and $t$ to the inertial frame. The blue arrow $\Matrix{A_{t-\delta}}{F}{A_t}$ illustrates the \emph{intra-body} transformation in the temporal interval $\delta$. Intra-body Galilean frames are independent of the choice of inertial frame $\{I_0\}$, and are a key concept for the geometric spatial-temporal estimation framework presented in this paper.}
    \label{fig:2_inertial}
\end{figure}

Consider an inertial frame at zero time $\{I_0\} \equiv \eyen{5}$.
Assume a positive time delay ${\delta \in \R^\plus}$, and let $\{A_t\}$ and $\{A_{t-\delta}\}$ denote the coordinate frames attached to a rigid body $A$ at times $t$ and $t-\delta$ respectively, as shown in \figref{2_inertial}.
The coordinates of $\{A_t\}$ with respect to $\{I_0\}$ are described by the Galilean frame
\begin{align}
\Matrix{I_0}{F}{A_t} =
    \begin{bmatrix} 
        \Matrix{I_0}{R}{A_t} & \Vector{I_0}{v}{A_t} & \Vector{I_0}{p}{A_t} \\
        \zeronm{1}{3} & 1 & ^{I_0}t_{A_t} \\
        \zeronm{1}{3} & 0 & 1 
    \end{bmatrix},
\end{align}
where ${\Matrix{I_0}{R}{A_t} \in \mathcal{SO}(3)}$ encodes the orientation, ${\Vector{I_0}{v}{A_t} \in \mathbb{R}^3}$ the inertial velocity, ${\Vector{I_0}{p}{A_t} \in \mathbb{R}^3}$ the position, and ${^{I_0}t_{A_t} = t}$ the time offset.
The same structure applies to the delayed time $t-\delta$ by replacing the time index accordingly.

We distinguish \emph{intra-body} transformations, defined for the same rigid-body frame at different times, from \emph{cross-body} transformations, describing the spatio-temporal relation between inertial frames attached to different rigid bodies.
Define the \emph{intra-body} Galilean frame ${\Matrix{A_{t-\delta}}{F}{A_t} = \Matrix{I_0}{F}{A_{t-\delta}}^{\!\!\!\!\!\!\!-1}\Matrix{I_0}{F}{A_t}}$ as
\begin{align}
    \Matrix{A_{t-\delta}}{F}{A_t} =
    \begin{bmatrix} 
        \Matrix{A_{t-\delta}}{R}{A_t} & \Vector{A_{t-\delta}}{v}{A_t} & \Vector{A_{t-\delta}}{p}{A_t} \\
        \zeronm{1}{3} & 1 & \delta \\
        \zeronm{1}{3} & 0 & 1 
    \end{bmatrix} ,
    \label{eq:intra_frame}
\end{align}
encoding the relative orientation, velocity, position, and time of $\{A\}$ over the time interval $\delta$.
Note that, by construction, the temporal offset ${^{A_{t-\delta}}t_{A_t} = t - (t - \delta) = \delta}$ is the \emph{time delay}.

\begin{figure}[t]
    \centering
    \includegraphics[width=0.6\linewidth]{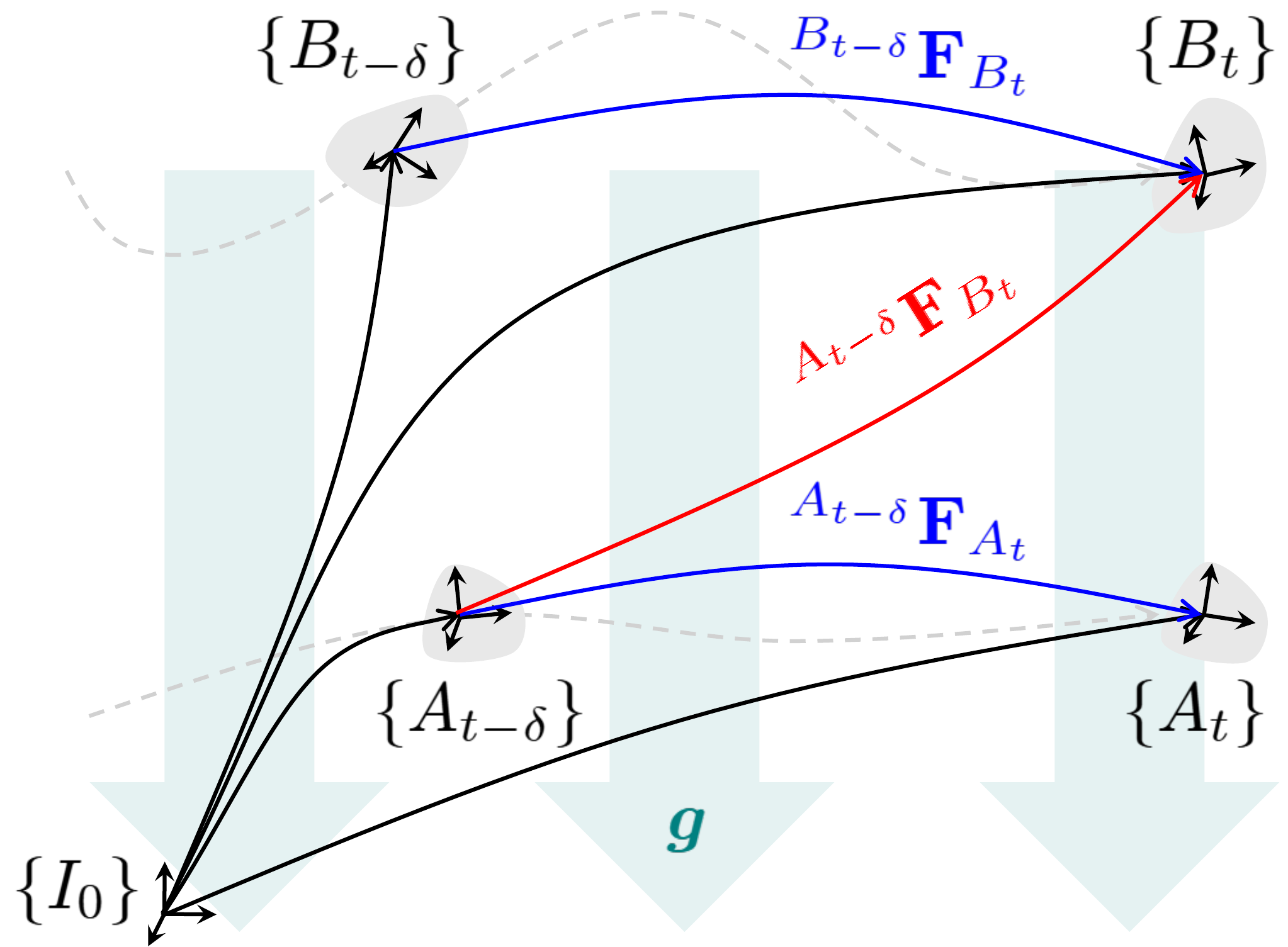}\\
    \caption{Galilean transformations between two rigid bodies $A$ and $B$ in a gravity field. This setup extends the single-body case of \figref{2_inertial} to account for the relative motion between two different rigid bodies. The blue arrows represent the \emph{intra-body} transformations $\Matrix{A_{t-\delta}}{F}{A_t}$ and $\Matrix{B_{t-\delta}}{F}{B_t}$ over the temporal interval $\delta$. The red arrow illustrates the \emph{cross-body} transformation $\Matrix{A_{t-\delta}}{F}{B_t}$, mapping coordinates from body $B$ at current time $t$ to body $A$ at an earlier time $t-\delta$. The key idea is to filter for the cross-body Galilean frame, enabling joint estimation of spatial navigation states and time delay $\delta$.}
    \label{fig:2_body}
\end{figure}

Consider a second rigid body $B$, as shown in \figref{2_body}.
The intra-body transformation ${\Matrix{B_{t-\delta}}{F}{B_t} = \Matrix{I_0}{F}{B_{t-\delta}}^{\!\!\!\!\!\!\!-1}\Matrix{I_0}{F}{B_t}}$ is defined as in~\eqref{eq:intra_frame} by replacing the frame indices accordingly.
Define the \emph{cross-body} Galilean frame ${\Matrix{A_{t-\delta}}{F}{B_t} = \Matrix{I_0}{F}{A_{t-\delta}}^{\!\!\!\!\!\!\!-1}\Matrix{I_0}{F}{B_t}}$ as
\begin{align}
    \Matrix{A_{t-\delta}}{F}{B_t} =
    \begin{bmatrix} 
        \Matrix{A_{t-\delta}}{R}{B_t} & \Vector{A_{t-\delta}}{v}{B_t} & \Vector{A_{t-\delta}}{p}{B_t} \\
        \zeronm{1}{3} & 1 & \delta \\
        \zeronm{1}{3} & 0 & 1 
    \end{bmatrix} ,
    \label{eq:cross_frame}
\end{align}
with $^{A_{t-\delta}}t_{B_t} = t - (t - \delta) = \delta$.
Leveraging Galilean symmetry ensures that the dynamics of each body are interpreted consistently with respect to an inertial reference frame, and provides a principled framework for representing coupled spatial and temporal transformations.
Designing a filter for cross-body Galilean frames enables joint estimation of spatial and temporal states in multi-body INS with delays.

\subsection{Kinematics of Galilean Frames}
\label{sec:kinematics}
We consider the continuous-time kinematics of Galilean frames under inertial excitation.
Given IMU measurements of angular velocity $\Vector{}{\omega}{} \in \Rn{3}$ and linear acceleration $\Vector{}{a}{} \in \Rn{3}$, define the input matrix $\Matrix{}{U}{}(\Vector{}{\omega}{}, \Vector{}{a}{}) \in \Rnm{5}{5}$ and a constant $\Matrix{}{N}{} \in \Rnm{5}{5}$ as
\begin{align}
    &\!\!\!\!\!\!\Matrix{}{U}{}(\Vector{}{\omega}{}, \Vector{}{a}{})\coloneqq
        \begin{bmatrix}
            \Vector{}{\omega}{}^\wedge & \Vector{}{a}{} & \zeronm{3}{1} \\ 
            \zeronm{2}{3} & \zeronm{2}{1} & \zeronm{2}{1}\\
        \end{bmatrix} ,
    &&\!\!\!\Matrix{}{N}{}\coloneqq
        \begin{bmatrix}
            \zeronm{3}{4} & \zeronm{3}{1} \\ 
            \zeronm{1}{4} & 1\\
            \zeronm{1}{4} & 0
        \end{bmatrix} .
        \label{eq:UN}
\end{align}
Let ${\Matrix{}{U}{A_t} = \Matrix{}{U}{}(\Vector{}{\omega}{A_t}, \Vector{}{a}{A_t})}$ and ${\Matrix{}{U}{A_{t-\delta}} = \Matrix{}{U}{}(\Vector{}{\omega}{A_{t-\delta}}, \Vector{}{a}{A_{t-\delta}})}$ be the inertial input matrices for rigid body $A$ at time $t$ and $t-\delta$.
The corresponding Galilean frames evolve according to
\begin{align}
    \dotMatrix{I_0}{F}{A_{t-\delta}} &= \Matrix{I_0}{F}{A_{t-\delta}} (\Matrix{}{U}{A_{t-\delta}} + \Matrix{}{N}{}) , \\
    \dotMatrix{I_0}{F}{A_t} &= \Matrix{I_0}{F}{A_t} (\Matrix{}{U}{A_t} + \Matrix{}{N}{}) .
\end{align}
Note that the role of $\Matrix{}{N}{}$ is to model the \emph{time} evolution ${\dot{t} = 1}$.
The kinematics of the intra-body frame $\Matrix{A_{t-\delta}}{F}{A_t}$
follow as
\begin{align}
    &\dotMatrix{A_{t-\delta}}{F}{A_t} = \frac{d}{dt}\left(\Matrix{I_0}{F}{A_{t-\delta}}^{\!\!\!\!\!\!\!-1}\Matrix{I_0}{F}{A_t}\right) \\
    &= -\Matrix{I_0}{F}{A_{t-\delta}}^{\!\!\!\!\!\!\!-1} \dotMatrix{I_0}{F}{A_{t-\delta}} \Matrix{I_0}{F}{A_{t-\delta}}^{\!\!\!\!\!\!\!-1} \Matrix{I_0}{F}{A_t} + \Matrix{I_0}{F}{A_{t-\delta}}^{\!\!\!\!\!\!\!-1}\dotMatrix{I_0}{F}{A_t} \\
    &= -(\Matrix{}{U}{A_{t-\delta}} + \Matrix{}{N}{})\Matrix{A_{t-\delta}}{F}{A_t} + \Matrix{A_{t-\delta}}{F}{A_t}(\Matrix{}{U}{A_t} + \Matrix{}{N}{}) ,
    \label{eq:intra_frame_kinematic}
\end{align}
and the same goes for $\dotMatrix{B_{t-\delta}}{F}{B_t}$ by replacing frame indices.
An analogous computation yields the kinematics of the cross-body frame $\Matrix{A_{t-\delta}}{F}{B_t}$, even though its kinematics depend on inputs at two different times,  $t-\delta$ and $t$:
\begin{align}
    &\dotMatrix{A_{t-\delta}}{F}{B_t} = \frac{d}{dt}\left(\Matrix{I_0}{F}{A_{t-\delta}}^{\!\!\!\!\!\!\!-1}\Matrix{I_0}{F}{B_t}\right) \\
    &= -\Matrix{I_0}{F}{A_{t-\delta}}^{\!\!\!\!\!\!\!-1} \dotMatrix{I_0}{F}{A_{t-\delta}} \Matrix{I_0}{F}{A_{t-\delta}}^{\!\!\!\!\!\!\!-1} \Matrix{I_0}{F}{B_t} + \Matrix{I_0}{F}{A_{t-\delta}}^{\!\!\!\!\!\!\!-1}\dotMatrix{I_0}{F}{B_t} \\
    &= -(\Matrix{}{U}{A_{t-\delta}} + \Matrix{}{N}{})\Matrix{A_{t-\delta}}{F}{B_t} + \Matrix{A_{t-\delta}}{F}{B_t}(\Matrix{}{U}{B_t} + \Matrix{}{N}{}) ,
    \label{eq:cross_frame_kinematic}
\end{align}
where ${\Matrix{}{U}{B_t} = \Matrix{}{U}{}(\Vector{}{\omega}{B_t}, \Vector{}{a}{B_t})}$.

\subsection{Application to Spatio-Temporal State Estimation}
\label{sec:state_estimation}
\begin{figure}[t]
    \centering
    \includegraphics[width=0.6\linewidth]{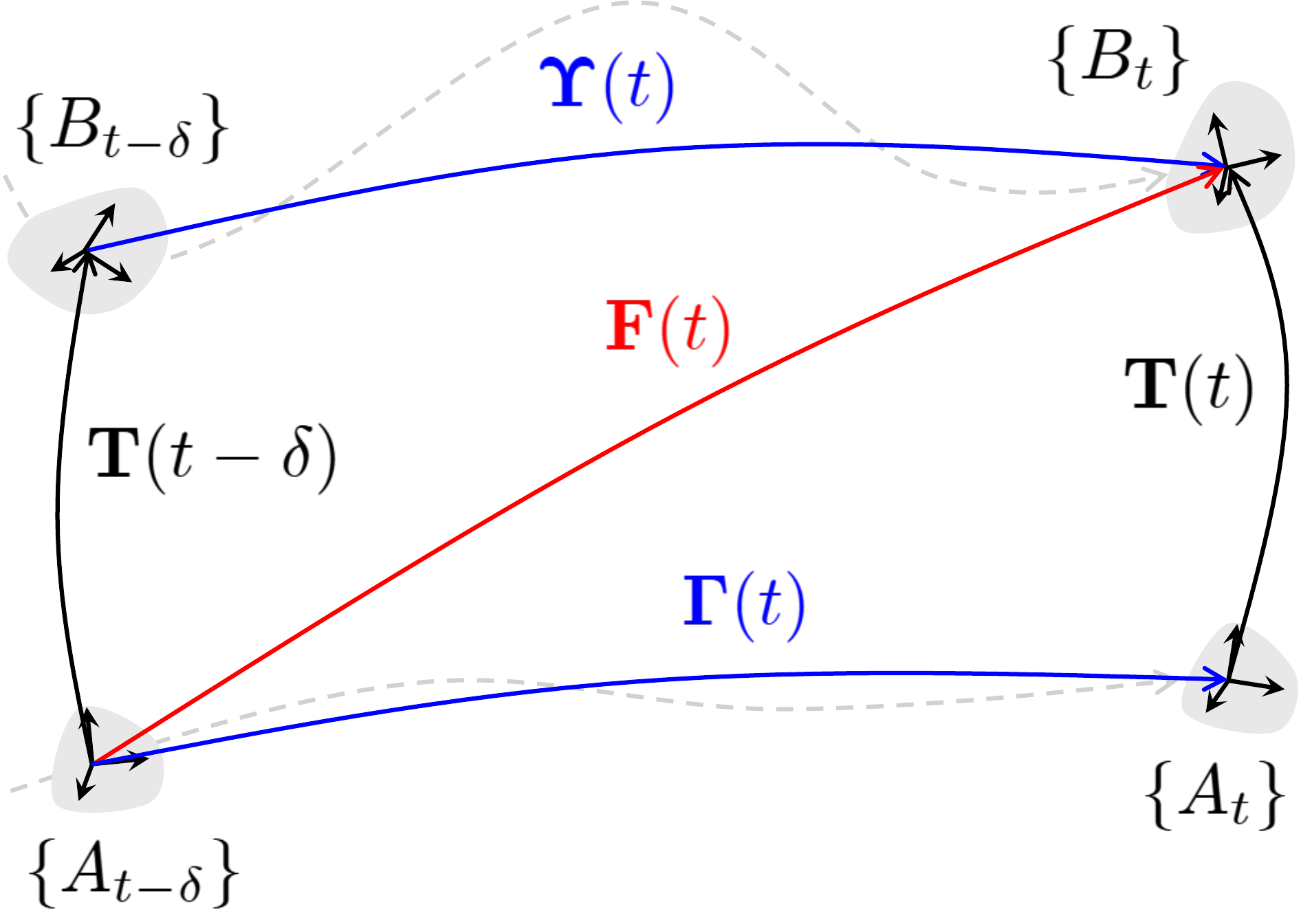}\\
    \caption{Representation of the time-varying Galilean frames between bodies $A$ and $B$ illustrated in \figref{2_body}. The complex frame-dependent notation is replaced with compact notation: $\boldsymbol{\Gamma}(t)$ and $\boldsymbol{\Upsilon}(t)$ denote the \emph{intra-body} transformations (blue) for bodies $A$ and $B$ respectively, while $\mathbf{F}(t)$ denotes the \emph{cross-body} transformation (red) between the two rigid bodies. $\mathbf{T}(t)$ and $\mathbf{T}(t-\delta)$ denote \emph{isochronous} Galilean frames, equivalent to extended poses, which encode the spatial transformation between inertial frames at the same time instant.}
    \label{fig:2_frames}
\end{figure}

We provide insights on how to use the proposed framework to derive a simple state observer for the relative motion and time delay of two rigid bodies with delayed relative pose measurements.
The notation used is illustrated in \figref{2_frames}.

Define the system's state ${\Matrix{}{F}{}(t) \in \mathcal{G}al(3)}$ as the cross-body Galilean frame~\eqref{eq:cross_frame}, with kinematics given by~\eqref{eq:cross_frame_kinematic}.
The state observer is $\hat{F} = \fourel{\hat{R}}{\hat{v}}{\hat{p}}{\hat{\delta}} \in \G{3}$, with kinematics
\begin{equation}
    \dot{\hat{F}} = -U_{A_{t-\delta}} \hat{F} + \hat{F} U_{B_t} + \hat{F} N - N \hat{F} + \Delta \hat{F},
    \label{eq:observer_kinematics}
\end{equation}
where ${\Delta \in \g{3}}$ is an innovation term driving ${\hat{F}(t) \rightarrow \Matrix{}{F}{}(t)}$.
The time-varying inertial inputs ${U_{A_{t-\delta}}, U_{B_t} \in \g{3}}$ and the constant ${N \in \g{3}}$ are defined according to~\eqref{eq:UN} and represented as elements of the Lie algebra of $\G{3}$.
Note that the observer kinematics~\eqref{eq:observer_kinematics} is group-affine and preserves the Lie group structure of the state space.
The extended poses between the two rigid bodies ${\Matrix{}{T}{}(t-\delta), \Matrix{}{T}{}(t) \in \mathcal{G}al(3)}$ at times ${t-\delta}$ and $t$ are represented as \emph{isochronous} Galilean frames, i.e., their temporal component is zero, and can be modeled as
\begin{align}
    &\Matrix{}{T}{}(t-\delta) = \Matrix{}{F}{}(t) \Matrix{}{\Upsilon}{}(t)^{-1},
    &&\Matrix{}{T}{}(t) = \Matrix{}{\Gamma}{}(t)^{-1} \Matrix{}{F}{}(t) .
    \label{eq:delayed_position}
\end{align}
Note that modeling a delayed pose measurement or retrieving the current relative pose from the state requires the two intra-body transformations ${\Matrix{}{\Gamma}{}(t), \Matrix{}{\Upsilon}{}(t) \in \mathcal{G}al(3)}$.
These Galilean frames depend on the time delay $\delta$ and buffered IMU measurements, and are termed \emph{preintegration matrices}.
The idea of using Galilean symmetry to provide a compact formulation of IMU preintegration has been recently introduced by~\citet{Delama2025EquivariantApproach}.
These terms are treated as known functions of the time delay, denoted as ${\Gamma(\delta), \Upsilon(\delta) \in \G{3}}$.
For a fixed delay $\delta$, the preintegration matrices can be recursively computed from buffered IMU data.
\emph{Without loss of generality, the definition is given for $\Upsilon(\delta)$, with $\Gamma(\delta)$ defined analogously by replacing the frame indices with $A$.}
Given buffered IMU measurements ${\{U_{B_{t-\delta}}, \cdots, U_{B_t}\}}$, the preintegration matrix is
\begin{equation}
    \Upsilon(\delta) = \prod_{k=(t - \delta)}^t \exp((U_{B_k} + N)\dt) ,
    \label{eq:upsilon_gal3}
\end{equation}
where ${\dt \in \R}$ is the IMU time step.
Recomputing the preintegration matrix from scratch each time the time delay changes during estimation is possible, but highly inefficient.
Alternatively, it is possible to compute it once and propagate it as long as the delay converges to a steady value.
The discrete-time kinematics of $\Upsilon$ is obtained by discretizing~\eqref{eq:intra_frame_kinematic} as
\begin{equation}
    \Upsilon_{i+1} = -\exp((U_{B_{t-\delta}} + N)\dt) \Upsilon_i \exp((U_{B_t} + N)\dt) ,
    \label{eq:upsilon_kinematic}
\end{equation}
where the subscript $i$ denotes the index of the observer iterative loop.
However, this propagation introduces drift due to noise in the IMU measurements, and if the delay changes, the preintegration matrix must be recomputed.
In practice, a more robust approach is to keep a \emph{sliding buffer} of preintegrated IMU matrices over a fixed time window, e.g., 0.5\,s.
At each iteration $i$, a new identity element is appended, and the oldest entry is removed if the buffer exceeds its maximum length.
Then, all entries in the buffer are propagated in batch as
\begin{equation}
    \Upsilon_k \gets \Upsilon_k \exp((U_{B_t} + N)\dt) ,
    \label{eq:upsilon_update}
\end{equation}
where the subscript $k$ identifies the preintegration matrix at the $k$‑th position in the sliding window.
The function $\Upsilon(\delta)$ then retrieves the preintegration matrix corresponding to the requested delay, using interpolation if needed.
This strategy allows efficient handling of variable delays without recomputing preintegration or introducing drift, increasing memory usage, but keeping the computational cost constant.
Using the preintegration matrices, the observer compares the \emph{measured} and \emph{predicted} relative poses to construct a residual
\begin{equation}
    \Vector{}{r}{} = f(\Matrix{}{T}{m}(t-\delta), \hat{F}) \in \Rn{m},
\end{equation}
which, for an extended pose measurement, can take the form
\begin{equation}
    \Vector{}{r}{} = \log(\Matrix{}{T}{m}(t-\delta) \Upsilon(\hat{\delta}) \hat{F}^{-1})^\vee,
\end{equation}
where ${\log: \G{3} \to \g{3}}$ maps the group error to the Lie algebra.
The residual $\Vector{}{r}{}$ is used to drive the innovation $\Delta$, providing a feedback correction that updates the observer toward the true state.
This correction is computed in the Lie algebra using a gain matrix as ${\Delta = (\Matrix{}{K}{} \Vector{}{r}{})^\wedge}$.
Finally, the relative extended pose estimate at the current time is retrieved:
\begin{equation}
    \hatMatrix{}{T}{}(t) = \Gamma(\hat{\delta})^{-1} \hat{F} .
\end{equation}

\section{Equivariant Filter for Delayed GNSS–INS}
\label{sec:filter}
This section derives a \emph{discrete-time} EqF for biased-INS with delayed GNSS measurements using the geometric framework introduced in~\secref{problem}.
\emph{The filter is designed to jointly estimate the system’s position, velocity, and attitude, as well as the unknown GNSS time delay and IMU biases.}
The system is formalized in~\secref{biased_ins} and the linearized error dynamics for the propagation are derived in~\secref{lifted_system}.
Finally, the update for delayed GNSS position measurements is formulated in~\secref{gnss_update}.
A practical implementation is outlined in~\algoref{eqf}.

\subsection{Biased-INS with Delayed Position Measurements}
\label{sec:biased_ins}
Consider a mobile robot equipped with an IMU that delivers \emph{biased} angular velocity and linear acceleration measurements, and a GNSS sensor providing delayed 3D position measurements affected by an \emph{unknown} time delay ${\delta \in \R^\plus}$.
The system is modeled using the geometric framework in~\figref{2_frames}.  
In this scenario, the coordinate frame $\{A\}$ is fixed to the Earth rather than attached to a second rigid body, and is subject to a constant input representing the Earth's angular velocity and a linear acceleration counteracting gravity.
Consider the system
\begin{equation}
    \dot\xi = f(\xi, u) \quad\Longleftrightarrow\quad
    \left\{\begin{aligned}
        \dot{\Matrix{}{F}{}} &= -\gn^\wedge\Matrix{}{F}{} + \Matrix{}{F}{}({\wn} - \Vector{}{b}{})^\wedge \\
        \dot{\Vector{}{b}{}} &= \Vector{}{\tau}{}
    \end{aligned}\right. ,
    \label{eq:systemCt}
\end{equation}
where the state is denoted as ${\xi = \twoel{\Matrix{}{F}{}}{\Vector{}{b}{}} \in \mathcal{M} \coloneqq \mathcal{G}al(3) \times \Rn{10}}$ and the input is represented as ${u = \twoel{\wn}{\Vector{}{\tau}{}} \in \mathbb{L} \subset \Rn{20}}$.
In particular, ${\Matrix{}{F}{} = \fourel{\Matrix{}{R}{}}{\Vector{}{v}{}}{\Vector{}{p}{}}{\delta} \in \mathcal{G}al(3)}$ denotes the \emph{core states}, including spatial coordinates and the temporal offset $\delta$, and ${\Vector{}{b}{} = \fourel{\Vector{}{b}{\omega}}{\Vector{}{b}{a}}{\Vector{}{b}{\nu}}{b_{\rho}} \in \Rn{10}}$ denotes the \emph{bias states}, including the IMU biases ${\Vector{}{b}{\omega}, \Vector{}{b}{a} \in \Rn{3}}$ and two \emph{virtual} biases ${\Vector{}{b}{\nu} \in \Rn{3}}$ and ${b_{\rho} \in \R}$ introduced by~\citet{Delama2025EquivariantApproach} to model the underlying symmetry.
${\tauInp{}{} = \fourel{\tauInp{}{\omega}}{\tauInp{}{a}}{\tauInp{}{\nu}}{\tau_\rho} \in \Rn{10}}$ models the bias random-walk.
Finally, ${\gn = \fourel{\Vector{}{\omega}{E}}{-\Vector{}{g}{}}{\zeronm{3}{1}}{1} \in \Rn{10}}$ is a \emph{constant} vector capturing the effect of the Earth's rotation ${\Vector{}{\omega}{E} \in \Rn{3}}$ and gravity, and ${\wn = \fourel{\Vector{}{\omega}{}}{\Vector{}{a}{}}{\zeronm{3}{1}}{1} \in \Rn{10}}$ is the time-varying extended \emph{biased inertial input} at current time.
The wedge operator ${(\cdot)^\wedge : \Rn{10} \to \g{3} \subset \Rnm{5}{5}}$ is used to map vectors to matrices.
Note that the terms ${\gn^\wedge = \Matrix{}{U}{}(\Vector{}{\omega}{E}, -\Vector{}{g}{}) +\Matrix{}{N}{}}$ and ${\wn^\wedge = \Matrix{}{U}{}(\Vector{}{\omega}{}, \Vector{}{a}{}) + \Matrix{}{N}{}}$ derive from~\eqref{eq:UN}, and the system kinematics depend only on the input at the \emph{current IMU time}.

By assuming a constant input ${u_i \in \mathbb{L}}$ between consecutive timestamps $t_i$ and $t_{i+1}$, and defining the IMU time step ${\dt = t_{i+1} - t_i}$, the exact discretization of~\eqref{eq:systemCt} yields
\begin{equation}
    \left\{\begin{aligned}
    \Matrix{}{F}{i+1} &= \exp(-\gn^\wedge\dt)\Matrix{}{F}{i}\exp(({\wn}_i - \Vector{}{b}{i})^\wedge\dt) \\
    \Vector{}{b}{i+1} &= \Vector{}{b}{i} + \Vector{}{\tau}{i}\dt 
    \end{aligned}\right. ,
    \label{eq:systemDt}
\end{equation}
where the subscript $i$ denotes the index of the iterative loop.

\subsection{Lifted System and Error Dynamics}
\label{sec:lifted_system}
\begin{algorithm}[t]
\caption{Equivariant Filter with Online Delay Estimation}
\begin{flushleft}
    \textbf{Parameters:} ${\mathring{\xi} = \twoel{\eyen{5}}{\zeronm{10}{1}}}$, $\xi_{init}$, ${\mathbf{\Sigma}_{init}}$, $\mathbf{Q}$, $\mathbf{R}$, $\Vector{}{p}{0}$ \\
    \textbf{Initialization:} ${\hat{X} = \phi^{-1}_{\mathring{\xi}}(\xi_{init})}$, ${\hat{\mathbf{\Sigma}} = \mathbf{\Sigma}_{init}}$ \\
    \textbf{Measurements:} IMU $\rightarrow \twoel{\omegaInp{}{}}{\aInp{}{}}$, GNSS $\rightarrow \Vector{}{p}{}(t-\delta)$ \\
    \textbf{Output:} $\hat{X}$, $\hatMatrix{}{\Sigma}{}$, ${\hatMatrix{}{T}{}(t) = \Gamma(\hat{\delta})^{-1} \hat{X}_F}$ 
\end{flushleft}
\vspace{-3mm}
\hrulefill\\
\textsc{Propagation}: require an input meas. ${u = \twoel{\wn}{\tauInp{}{}}}$
\begin{algorithmic}[1]
    \State ${\hat{\mathbf{\Sigma}} \gets \mathbf{A} \hat{\mathbf{\Sigma}} \mathbf{A}^\top + \frac{1}{\dt} \mathbf{B} \mathbf{Q} \mathbf{B}^\top }$ \hfill \eqref{eq:Adt}~\eqref{eq:Bdt}
    \State ${\hat{X} \gets \hat{X} \Lambda(\phi(\hat{X}, \mathring{\xi}), \tilde{u})}$ \hfill \eqref{eq:lifted_system}
    \State $\Upsilon_k \gets \Upsilon_k \exp((\wn - \hatbias{}{})^\wedge \dt) \ \forall k$ \hfill \eqref{eq:upsilon_update}
\end{algorithmic}
\textsc{Update}: require an output meas. ${y_m = \Vector{}{p}{}(t-\delta)}$
\begin{algorithmic}[1]
    \State ${\Vector{}{r}{} \gets y_m - h(\phi(\hat{X}, \mathring{\xi}))}$ \hfill \eqref{eq:residual}
    \State ${\mathbf{K} \gets \hat{\mathbf{\Sigma}}\mathbf{C}^\top (\mathbf{C} \hat{\mathbf{\Sigma}} \mathbf{C}^\top + \mathbf{D} \mathbf{R} \mathbf{D}^\top)^{-1}}$ \hfill \eqref{eq:C}
    \State ${\Delta \gets \diff{\phi_{\mathring{\xi}}}^\dag \circ \diff{\vartheta^{-1}} \circ \mathbf{K} \Vector{}{r}{}}$ \hfill $^\star$
    \State ${\hat{\mathbf{\Sigma}} \gets (\eyen{20} - \mathbf{K}\mathbf{C})\hat{\mathbf{\Sigma}}}$
    \State ${\hat{\mathbf{\Sigma}} \gets \Jl{\Delta^\vee} \hat{\mathbf{\Sigma}} \Jl{\Delta^\vee}^\top}$ \hfill $^\star$$^\star$
    \State ${\hat{X} \gets \exp(\Delta) \hat{X}}$
\end{algorithmic}
\vspace{-1mm}
\hrulefill\\
\scriptsize{$^\star$Note that ${\diff{\phi_{\mathring{\xi}}}^\dag \circ \diff{\vartheta^{-1}}[\point] = (\point)^\wedge}$.} \hfill \scriptsize{$^\star$$^\star$This is the \emph{reset step} (\citet{GE2026106656}).}
\label{alg:eqf}
\end{algorithm}
The EqF is derived from a \emph{lifted system}, in which the original system is represented via a Lie group symmetry with an equivariant structure.
This formulation defines the state dynamics in a structured way and leverages the \emph{equivariant error} with its favorable linearization properties (\citet{vanGoor2022EquivariantEqF}).
Define the symmetry group ${\grpG \coloneqq \G{3} \ltimes \g{3}}$ called \emph{left-trivialized tangent group} of the Galilean group, proposed by~\citet{Delama2025EquivariantApproach}.
This group is the semi-direct product between $\G{3}$ and its Lie algebra.
For ${X = \twoel{A}{a} \in \grpG}$ and ${Y = \twoel{B}{b} \in \grpG}$, the group operation and the inverse are
\begin{align}
&\!\!\!\!XY = \twoel{AB}{a + \Adsym{A}{b}}, &&\inv{X} = \twoel{\inv{A}}{-\Adsym{\inv{A}}{a}}.
\end{align}
The lifted system state is ${X = \twoel{X_F}{X_b} \in \grpG}$, with kinematics
\begin{align}
    &X_{i+1} = X_i \Lambda\!\left(\phi(X_i, \mathring{\xi}), u_i\right) ,
    \label{eq:lifted_system}
\end{align}
where $\mathring{\xi} \in \mathcal{M}$ is an arbitrarily chosen \emph{state origin}.
The mapping ${\phi : \grpG \times \mathcal{M} \to \mathcal{M}}$ defines the \emph{state action} and is given by
\begin{equation}
    \phi(X, \xi) \coloneqq \twoel{\Matrix{}{F}{} X_F}{\AdMsym{X_F^{-1}}(\Vector{}{b}{} - X_b^\vee)} .
    \label{eq:phi}
\end{equation}
The mapping $\Lambda : \mathcal{M} \times \mathbb{L} \rightarrow \grpG$ is the \emph{discrete lift}, defined as
\begin{equation}
    \Lambda(\xi, u) \coloneqq \twoel{\Lift{F}(\xi, u)}{\Lift{b}(\xi, u)} ,
    \label{eq:lift}
\end{equation}
where $\Lift{F}\in \G{3}$ and $\Lift{b} \in \g{3}$ take the form
\begin{align}
    \Lift{F}(\xi, u) &= \exp(-\Adsym{\Matrix{}{F}{}^{-1}}{\gn^\wedge}\dt) \exp((\wn - \Vector{}{b}{})^\wedge\dt) , \\
    \Lift{b}(\xi, u) &= \Vector{}{b}{}^{\wedge} - \Adsym{\Lift{F}(\xi, u)}{\Vector{}{b}{}^\wedge + \Vector{}{\tau}{}^\wedge\dt} .
\end{align}

The EqF is formulated around the \emph{equivariant error}, which is intrinsically defined on the manifold $\mathcal{M}$ and evolves according to coordinate‑free dynamics.
This global error is linearized about the state origin $\mathring{\xi}$ to derive the filter propagation and update laws that drive filter convergence.
It is defined as
\begin{equation}
    e \coloneqq \phi(\hat{X}^{-1}, \xi) \in \mathcal{M} ,
\end{equation}
where ${\hat{X} = \twoel{\hat{X}_F}{\hat{X}_b} \in \grpG}$ denotes the \emph{state estimate} on the symmetry group, and ${\xi = \twoel{\Matrix{}{F}{}}{\Vector{}{b}{}} \in \mathcal{M}}$ represents the \emph{true state} on the manifold.
Define a local parametrization $\vartheta : \mathcal{M} \rightarrow \R^{20}$ for the error around $\mathring{\xi}$ using \emph{normal coordinates} as
\begin{equation}
    \Vector{}{\varepsilon}{} = \vartheta(e) = \log\!\left(\phi_{\mathring{\xi}}^{-1}(e)\right)^{\vee} = \log\!\left(\phi_{\mathring{\xi}}^{-1}\!\left(\phi(\hat{X}^{-1}, \xi)\right)\right)^{\vee} ,
\end{equation}
where ${\log(\point): \grpG \to \mathfrak{g}}$ denotes the logarithm of $\grpG$.

By assuming a \emph{noisy} input ${\tilde{u} = u + \eta \in \mathbb{L}}$ and fixing ${\mathring{\xi} = I}$, we derive the linearized error dynamics ${\Vector{}{\varepsilon}{i+1} \approx \Matrix{}{A}{} \Vector{}{\varepsilon}{i} + \Matrix{}{B}{} \Vector{}{\eta}{i}}$ about ${\Vector{}{\varepsilon}{} \!=\! \Vector{}{0}{}}$ and ${\Vector{}{\eta}{} \!=\! \Vector{}{0}{}}$.
The discrete-time \emph{state} matrix $\mathbf{A} \in \Rnm{20}{20}$ is
\begin{equation}
    \mathbf{A} = \begin{bmatrix}
        \Matrix{}{A}{1} & \Matrix{}{A}{1} \Jl{\ringwn \dt} \dt \\
        \zeronm{10}{10} & \Matrix{}{A}{2}
    \end{bmatrix} ,
    \label{eq:Adt}
\end{equation}
where ${\ringwn = \AdMsym{\hat{X}_F}{\wn} + \hat{X}_b^\vee}$, and $\Matrix{}{A}{1}, \Matrix{}{A}{2} \in \Rnm{10}{10}$ are
\begin{align}
    &\Matrix{}{A}{1} = \AdMsym{\exp(\gn^\wedge\dt)} , &&\Matrix{}{A}{2} = \Matrix{}{A}{1} \AdMsym{\exp(\ringwn^\wedge\dt)} .
\end{align}
The discrete-time \emph{input noise} matrix $\mathbf{B} \in \Rnm{20}{20}$ is 
\begin{equation}
    \mathbf{B} = \begin{bmatrix}
        -\Matrix{}{A}{1} \Jl{\ringwn\dt} \AdMsym{\hat{X}_F} \dt & \zeronm{10}{10} \\
        \zeronm{10}{10} & \Matrix{}{A}{2} \AdMsym{\hat{X}_F} \dt
    \end{bmatrix} .
    \label{eq:Bdt}
\end{equation}

\subsection{Delayed Position Measurement Update}
\label{sec:gnss_update}
A delayed position measurement ${y_m \in \Rn{3}}$ is modeled as
\begin{equation}
    \begin{pmatrix}
        y_m \\ 0 \\ 1
    \end{pmatrix} = \Matrix{}{T}{}(t-\delta) \barVector{}{y}{0} + \barVector{}{\mu}{} = \Matrix{}{F}{}(t)\Matrix{}{\Upsilon}{}(t)^{-1} \barVector{}{y}{0} + \barVector{}{\mu}{} ,
\end{equation}
with ${\barVector{}{y}{0} = \threeel{\Vector{}{p}{0}}{0}{1} \in \Rn{5}}$ and ${\barVector{}{\mu}{} = \twoel{\Vector{}{\mu}{}}{\zeronm{2}{1}} \in \Rn{5}}$, where ${\Vector{}{p}{0} \in \Rn{3}}$ is the GNSS receiver position in the IMU frame and the noise term ${\Vector{}{\mu}{} \sim \mathcal{N}(\zeronm{3}{1}, \Matrix{}{\Sigma}{\mu})}$ denotes the GNSS-position uncertainty.
Define the measurement function ${h : \mathcal{M} \to \Rn{3}}$ as
\begin{equation}
    h(\xi) = \mathcal{P}(\Matrix{}{F}{} \Upsilon(\delta)^{-1} \barVector{}{y}{0}) ,
    \label{eq:output}
\end{equation}
where ${\mathcal{P}(\Vector{}{x}{}) = \Vector{}{x}{1:3}}$ extracts the first three elements of a vector, and ${\Upsilon(\delta) \in \G{3}}$ is the IMU \emph{preintegration matrix} detailed in~\secref{state_estimation}, with the difference that here the buffered IMU data must be corrected using the corresponding bias estimate.

Upon receiving a GNSS position $y_m$, a residual ${\Vector{}{r}{} \in \Rn{3}}$ is defined by comparing the measurement with its prediction as
\begin{equation}
    \Vector{}{r}{} = y_m - h(\phi(\hat{X}, \mathring{\xi})) .
    \label{eq:residual}
\end{equation}
Linearizing ${\Vector{}{r}{} \approx \Matrix{}{C}{} \Vector{}{\varepsilon}{} + \Matrix{}{D}{} \Vector{}{\mu}{}}$ yields ${\Matrix{}{D}{} = \eyen{3}}$ and ${\Matrix{}{C}{} \in \Rnm{3}{20}}$ as
\begin{equation}
    \Matrix{}{C}{} = \begin{bmatrix}
        \Matrix{}{C}{1} & \zeronm{3}{3} & \eyen{3} & \Matrix{}{C}{2} & \zeronm{3}{10}
    \end{bmatrix} ,
    \label{eq:C}
\end{equation}
with ${\Matrix{}{C}{1}\in \Rnm{3}{3}}$ and ${\Matrix{}{C}{2}\in \Rnm{3}{1}}$ defined as
\begin{align}
    \Matrix{}{C}{1} &= -(\hat{R} p_{\Upsilon^{-1}} -\hat{v} \hat{\delta} + \hat{p})^\wedge , \\
    \Matrix{}{C}{2} &= \hat{v} - \hat{R} ( R_{\Upsilon^{-1}} \Vector{}{\omega}{t-\hat{\delta}}^\wedge\Vector{}{p}{0} + v_{\Upsilon^{-1}} ) ,
\end{align}
where ${\hat{X}_F = \fourel{\hat{R}}{\hat{v}}{\hat{p}}{\hat{\delta}} \in \G{3}}$ is the current state estimate, ${\Upsilon(\hat{\delta})^{-1} = \fourel{R_{\Upsilon^{-1}}}{v_{\Upsilon^{-1}}}{p_{\Upsilon^{-1}}}{-\hat{\delta}} \in \G{3}}$ is the inverse of the preintegration matrix~\eqref{eq:upsilon_gal3}, and ${\Vector{}{\omega}{t-\hat{\delta}} \in \Rn{3}}$ is the IMU angular velocity at time $t-\hat{\delta}$, retrieved from the buffer.
The discrete-time EqF is complete and summarized in \algoref{eqf}.
\section{Experiments and Results}
\label{sec:results}
The proposed EqF for GNSS-aided INS with delayed position measurements is validated using both real-world experiments and simulations, and compared against state-of-the-art EKF baselines.
We evaluate the performance using standard metrics for estimation accuracy and filter consistency.
The \emph{Absolute Rotation Error} (ARE), \emph{Absolute Position Error} (APE), and \emph{Absolute Velocity Error} (AVE) measure the differences between estimated and ground-truth states in the global frame.
These metrics are computed as the Euclidean norm of the corresponding error vectors, providing a scalar measure of the accuracy of rotation, position, and velocity estimation over time.
In addition, we utilize the \emph{Absolute Delay Error} (ADE) for time delays.
For a trajectory with $M$ time steps, let $(\Matrix{}{R}{i}, \Vector{}{p}{i}, \Vector{}{v}{i})$ denote the ground truth rotation, position, and velocity at time $t_i$, and $(\hatMatrix{}{R}{i}, \hatVector{}{p}{i}, \hatVector{}{v}{i})$ the corresponding filter estimates.
Let $\delta_i$ and $\hat{\delta}_i$ be the ground truth and estimated time delays.
The errors are computed for each timestamp as
\begin{align}
    &\mathrm{ARE}_i = \big\|\log(\Matrix{}{R}{i}^\top \hatMatrix{}{R}{i})^\vee \big\| \in \R,
    &&\mathrm{AVE}_i = \big\|\Vector{}{v}{i} - \hatVector{}{v}{i}\big\| \in \R,
    &&&\mathrm{APE}_i = \big\|\Vector{}{p}{i} - \hatVector{}{p}{i}\big\| \in \R,
    &&&&\mathrm{ADE}_i = |\delta_i - \hat{\delta}_i| \in \R.
\end{align}
To summarize these errors over the trajectory of length $M$, we report the \emph{Root Mean Square Error} (RMSE), defined as
\begin{equation}
    \mathrm{RMSE} = \sqrt{\frac{1}{M} \sum_{i=0}^{M-1} \mathrm{E}_i^2} \in \R,
\end{equation}
where $\mathrm{E}_i$ denotes either $\mathrm{ARE}_i$, $\mathrm{AVE}_i$, or $\mathrm{APE}_i$.
To evaluate filter consistency, we compute the \emph{Normalized Estimation Error Squared} (NEES), which measures the difference between estimation errors and the predicted covariance.
The full-state NEES is calculated for each timestamp as
\begin{equation}
    \mathrm{NEES}_i = \frac{1}{N} \boldsymbol{\varepsilon}_i^\top \mathbf{\Sigma}_i^{-1} \boldsymbol{\varepsilon}_i,
\end{equation}
where $\boldsymbol{\varepsilon}_i \in \Rn{N}$ is the state estimation error in local coordinates at time $t_i$, and $\mathbf{\Sigma}_i \in \Rnm{N}{N}$ is the corresponding covariance.

The trajectories used in both simulation and real-world experiments are designed to ensure sufficient excitation of the platform, including variations in linear acceleration and rotational motion.
Such excitation is desired to ensure that the time delay is observable, in accordance with classical observability results for the biased-INS.
In particular, degenerate cases such as the robot being stationary or moving at a constant velocity along a straight line make the delay unobservable.

\subsection{Real-World UAV Flights}
\begin{table*}[t]
\centering
\caption{Summary of real-world experiments with RMSEs for EKF and EqF under different time-delay settings.}
\label{tab:4_rmse}

\resizebox{\textwidth}{!}{%
\begin{tabular}{ccccc|cccc|cccc|cccc}
\toprule
\multicolumn{5}{c}{Experiment Metadata}
& \multicolumn{4}{c}{Rotation RMSE (deg)} 
& \multicolumn{4}{c}{Velocity RMSE (m/s)} 
& \multicolumn{4}{c}{Position RMSE (m)} \\
\midrule

\makecell{Dataset\\Sequence}
 & \makecell{Length\\(km)} 
 & \makecell{$v_{\mathrm{max}}$\\(m/s)}
 & \makecell{$a_{\mathrm{max}}$\\(m/s$^2$)}
 & \makecell{$\delta$\\(ms)}

 & \makecell{EKF\\(w/o delay)} 
 & \makecell{EKF\\(offline)} 
 & \makecell{EKF\\(online)} 
 & \makecell{EqF\\(online)}

 & \makecell{EKF\\(w/o delay)} 
 & \makecell{EKF\\(offline)} 
 & \makecell{EKF\\(online)} 
 & \makecell{EqF\\(online)}

 & \makecell{EKF\\(w/o delay)} 
 & \makecell{EKF\\(offline)} 
 & \makecell{EKF\\(online)} 
 & \makecell{EqF\\(online)} \\
\midrule

\texttt{striver}
& 2.10 & 34.82 & 51.31 & 120
& 5.54 & \textbf{3.49} & \underline{3.55} & 3.62
& 0.75 & 0.54 & \textbf{0.44} & \underline{0.45}
& 2.58 & \textbf{0.46} & \underline{0.58} & 0.63 \\

\texttt{hero-1}
& 4.15 & 38.15 & 18.94 & 90
& 3.22 & \underline{2.32} & 2.39 & \textbf{2.31}
& \underline{0.67} & 0.76 & 0.73 & \textbf{0.27}
& 2.62 & \underline{0.64} & 0.67 & \textbf{0.50} \\

\texttt{hero-2}
& 4.18 & 35.10 & 25.93 & 90
& 3.35 & \textbf{2.62} & \underline{2.76} & 2.84
& 0.76 & 0.46 & \underline{0.41} & \textbf{0.38}
& 2.51 & \textbf{0.61} & 0.66 & \underline{0.62} \\

\bottomrule
\end{tabular}%
}

\vspace{2mm}

{\fontsize{5.5}{6}\selectfont
\noindent
$v_{\scalebox{0.7}{max}}$: max speed.\quad
$a_{\scalebox{0.7}{max}}$: max acceleration.\quad
$\delta$: delay reference estimated offline.\quad
\emph{w/o delay}: delay fixed to 0 ms.\quad
\emph{offline}: delay fixed to $\delta$.\quad
\emph{online}: delay estimated online.\quad
\textbf{Best} results are bold; \underline{second-best} underlined.
}

\end{table*}
\begin{figure*}[t]
    \centering
    \subfloat[\texttt{striver}]{%
        \begin{minipage}[b]{0.32\textwidth}
            \includegraphics[width=\textwidth]{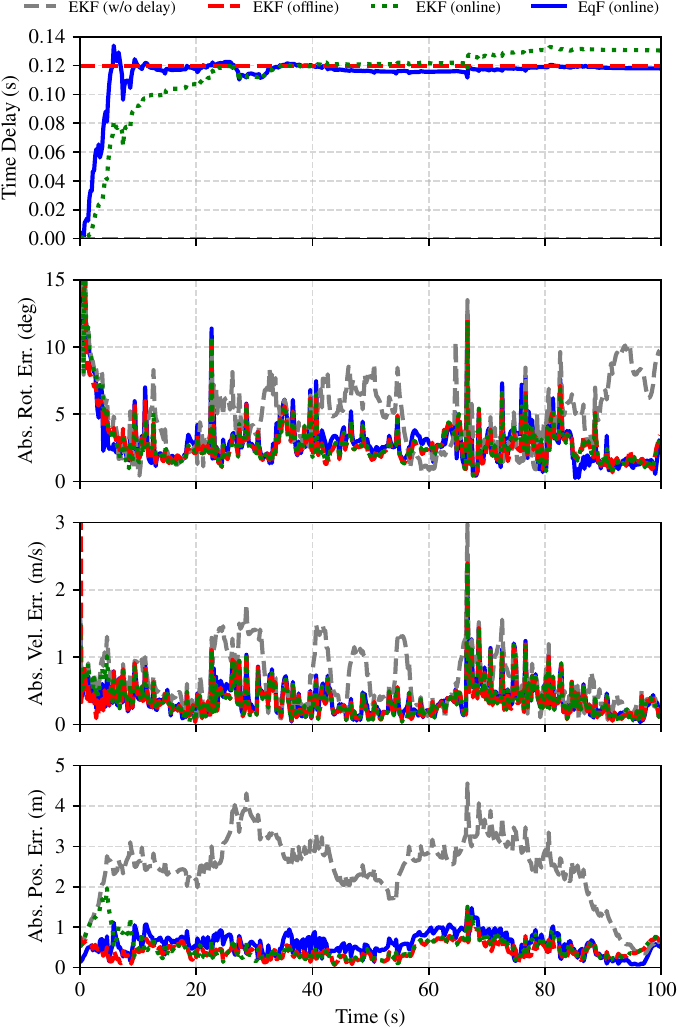}
        \end{minipage}%
    }
    \hfill
    \subfloat[\texttt{hero-1}]{%
        \begin{minipage}[b]{0.32\textwidth}
            \includegraphics[width=\textwidth]{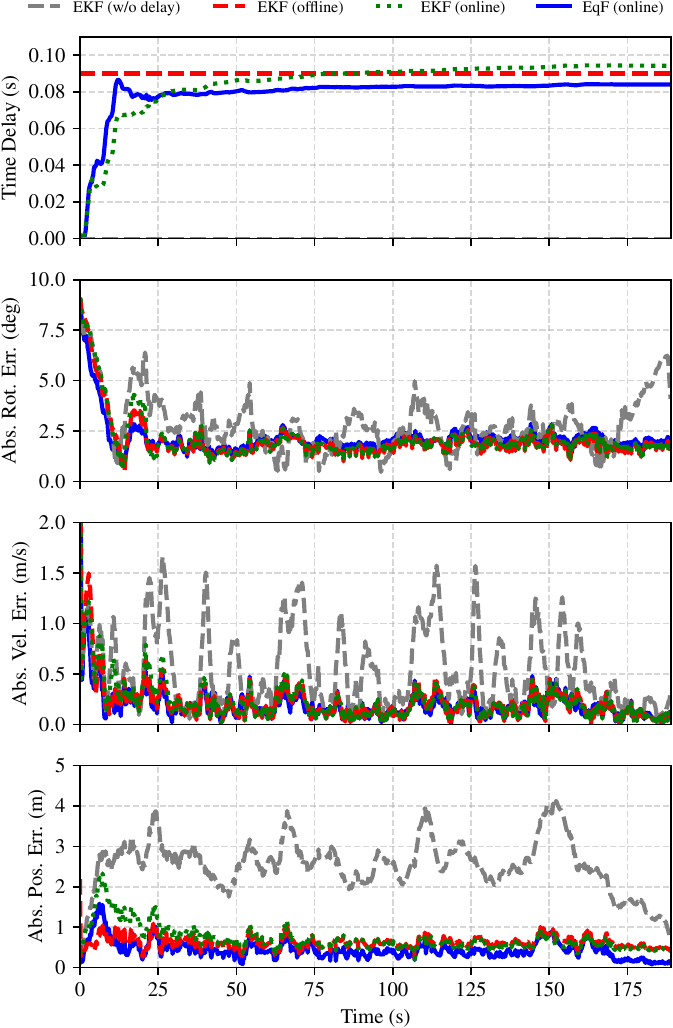}
        \end{minipage}%
    }
    \hfill
    \subfloat[\texttt{hero-2}]{%
        \begin{minipage}[b]{0.32\textwidth}
            \includegraphics[width=\textwidth]{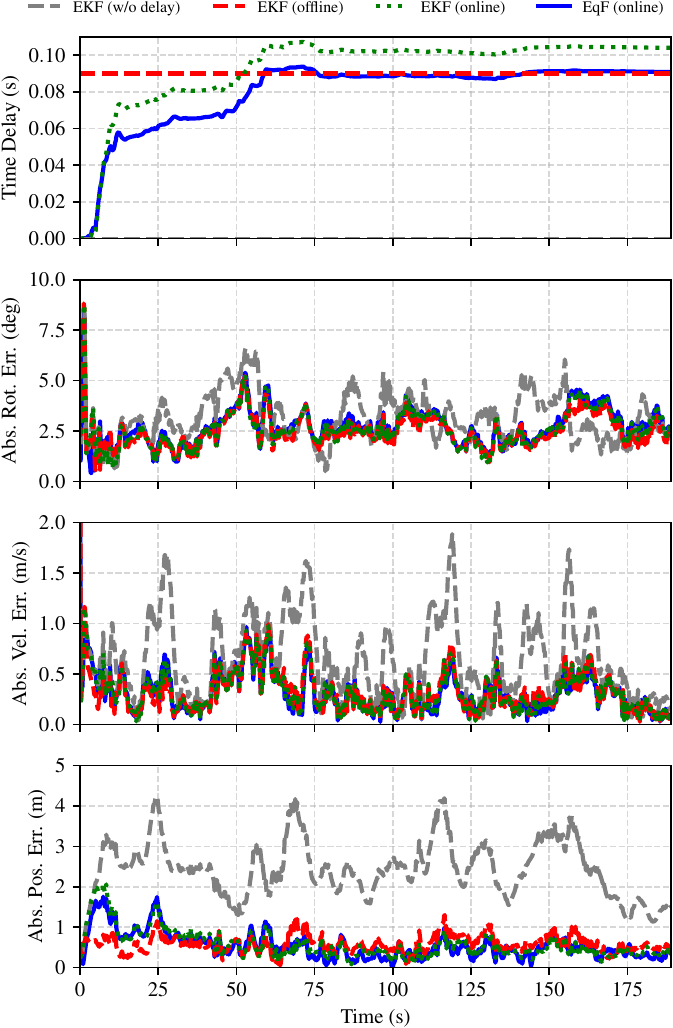}
        \end{minipage}%
    }
    \caption{Real-world comparison of the proposed EqF (solid blue) against three EKF baselines: a standard EKF without delay compensation (dashed grey), an EKF with offline delay estimate (dashed red), and an EKF with online time-delay estimation (dotted green). 
    The top row illustrates the \emph{online time-delay estimation}: (a) \texttt{striver} shows rapid, near-perfect convergence of our EqF to the 120\,ms delay reference, while EKF (online) shows a slower transient and a worse asymptotic; (b) \texttt{hero-1} shows a small $\sim$8\,ms residual bias of the EqF, suggesting the EqF estimate may be more accurate than the nominal 90\,ms delay, consistent with its superior performance in all other metrics in this specific dataset sequence; (c) \texttt{hero-2} shows a slower transient of both EqF and EKF (online), but ultimately the EqF converges to the 90\,ms delay reference. Overall, the proposed EqF consistently outperforms the EKF without delay compensation across all sequences, and matches or exceeds the accuracy of the EKF with offline and online delay compensation. Experiments metadata and RMSE values are reported in~\tabref{4_rmse}. For this magnitude of time delay, the real-world results are consistent with the simulation results illustrated in~\figref{4_mc_a}.}
    \label{fig:4_ardupilot}
\end{figure*}
\label{sec:rw}
The experimental validation is conducted using data collected from two fixed-wing UAV platforms, \texttt{striver} and \texttt{hero}, as illustrated in \figref{4_overview}.
The dataset consists of three flights: one performed by the \texttt{striver} and two by the \texttt{hero}.
In these scenarios, the GNSS measurements are subject to unknown ground-truth time delays.
A delay reference for each experiment is determined via an offline batch optimization that minimizes the state residuals over the full dataset, while the ground-truth trajectories are obtained from an on-board calibrated EKF that fuses measurements from additional sensors, including a magnetometer, barometer, and GNSS velocity.

We evaluate the proposed EqF by comparing its performance against three EKF baselines: \emph{EKF (w/o delay)}, i.e., a standard EKF without delay compensation, \emph{EKF (offline)}, i.e., an EKF utilizing a fixed offline-estimated delay compensation, and \emph{EKF (online)}, i.e., a state-of-the-art EKF with online delay estimation.
The latter is derived following the same design approach as~\citet{Kelly2021ASystems}, except that we use the extended pose formulation and the preintegration terms that are known from~\cite{Delama2025EquivariantApproach,Khosravian2016StateMeasurements}.
The EKF state is expressed as ${\xi_{EKF} = \threeel{\Matrix{}{T}{}}{\Vector{}{b}{}}{\delta} \in \mathcal{SE}_2(3) \times \Rn{6} \times \R}$, where $\Matrix{}{T}{}$ denotes the extended pose at the current time $t$ and the time delay $\delta$ is treated as an independent scalar.
Since the EKF does not exploit the Galilean frame structure that geometrically couples spatial and temporal states, the propagation differs from that of the EqF: there is no coupling between spatial and temporal state errors, and each component evolves independently in a standard Euclidean framework.
Moreover, the measurement function of the EKF is defined as ${h(\xi_{EKF}) = \mathcal{P}\big(\Gamma(\delta)\Matrix{}{T}{} \, \Upsilon(\delta)^{-1} \barVector{}{y}{0}\big)}$, which differs from~\eqref{eq:output}.  
It is worth noting that although we use preintegration terms $\Gamma$ and $\Upsilon$ introduced in~\secref{state_estimation} to reduce prediction error in the EKF, the residual linearization differs from that of the EqF, leading to different output matrices in the filter update.

The results presented in \figref{4_mc} show that the EqF successfully tracks time delays online across all three flight sequences, and the corresponding trajectory errors demonstrate that incorporating the delay significantly reduces state errors compared to the standard EKF without delay compensation.
Across the sequences, the EqF shows improved transient performance compared to the EKF with online delay estimation, and it achieves accuracy comparable to the EKF with offline-estimated delay compensation, while operating fully online.
The quantitative results, including experiment metadata and RMSE values, are summarized in \tabref{4_rmse}.
Note that the RMSE values are calculated over the entire trajectory, including the transient phase of the online delay estimation.
This introduces a slight disadvantage for EqF and EKF (online), since for these filters the delay is estimated in real time rather than being pre-calibrated as in EKF (offline).
Overall, the real-world experiments indicate that the proposed EqF can reliably operate fully online under unknown GNSS time delays.

\subsection{Simulation Experiments}
\label{sec:sim}
\begin{figure*}[b]
    \centering
    \subfloat[$\delta = 100\,\mathrm{ms}$]{%
        \begin{minipage}[b]{0.245\textwidth}
            \includegraphics[width=\textwidth]{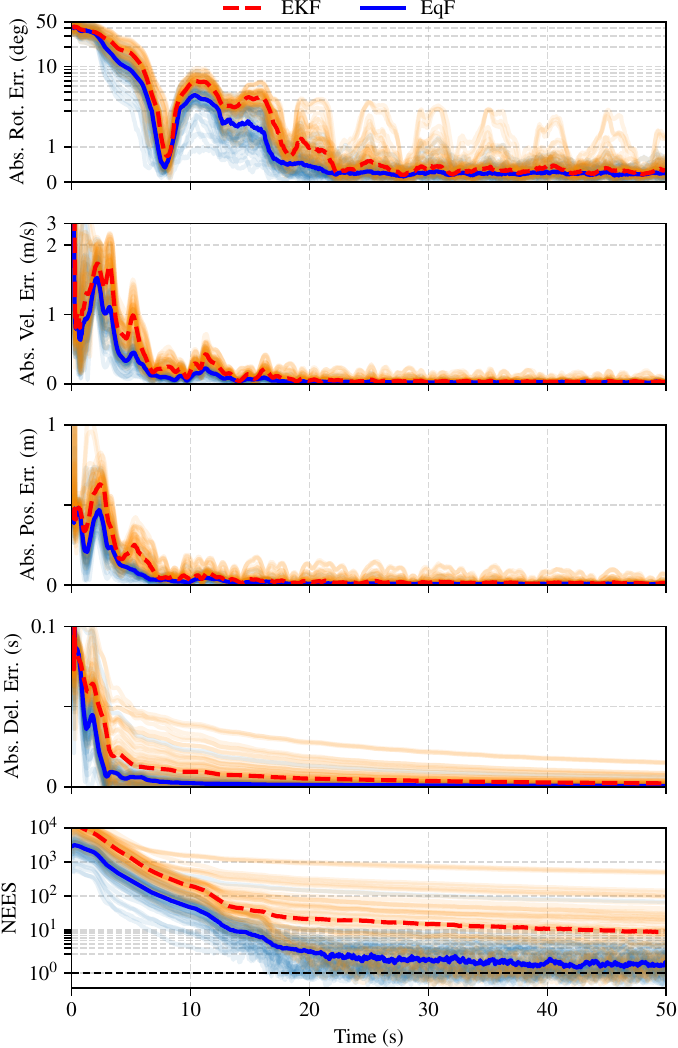}
        \end{minipage}%
        \label{fig:4_mc_a}
    }
    \hfill
    \subfloat[$\delta = 200\,\mathrm{ms}$]{%
        \begin{minipage}[b]{0.245\textwidth}
            \includegraphics[width=\textwidth]{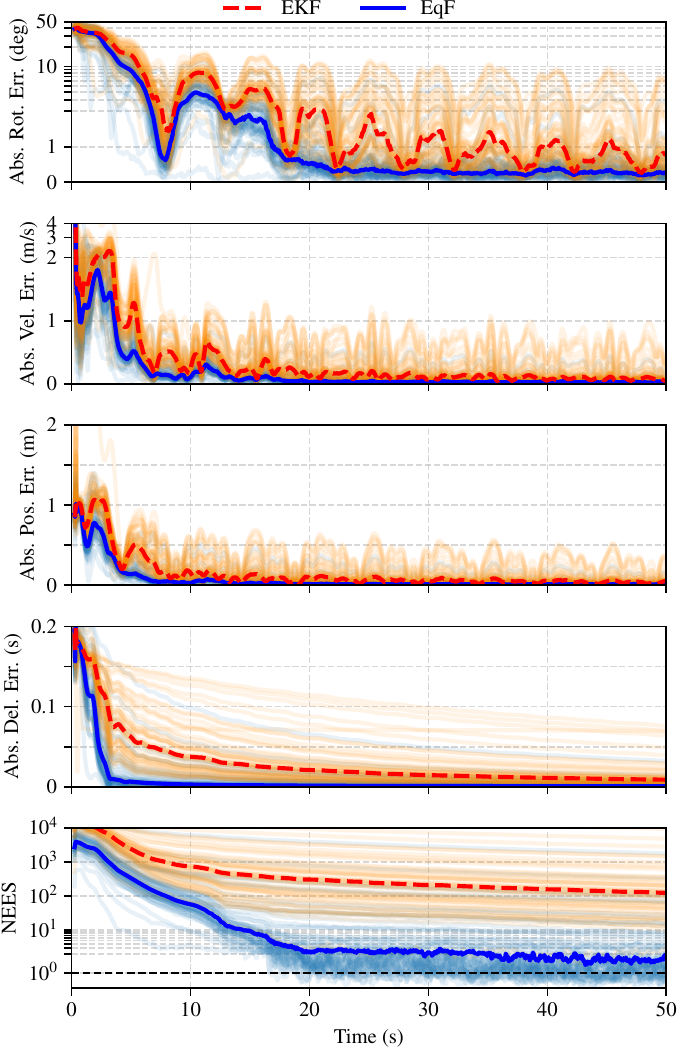}
        \end{minipage}%
    }
    \hfill
    \subfloat[$\delta = 300\,\mathrm{ms}$]{%
        \begin{minipage}[b]{0.245\textwidth}
            \includegraphics[width=\textwidth]{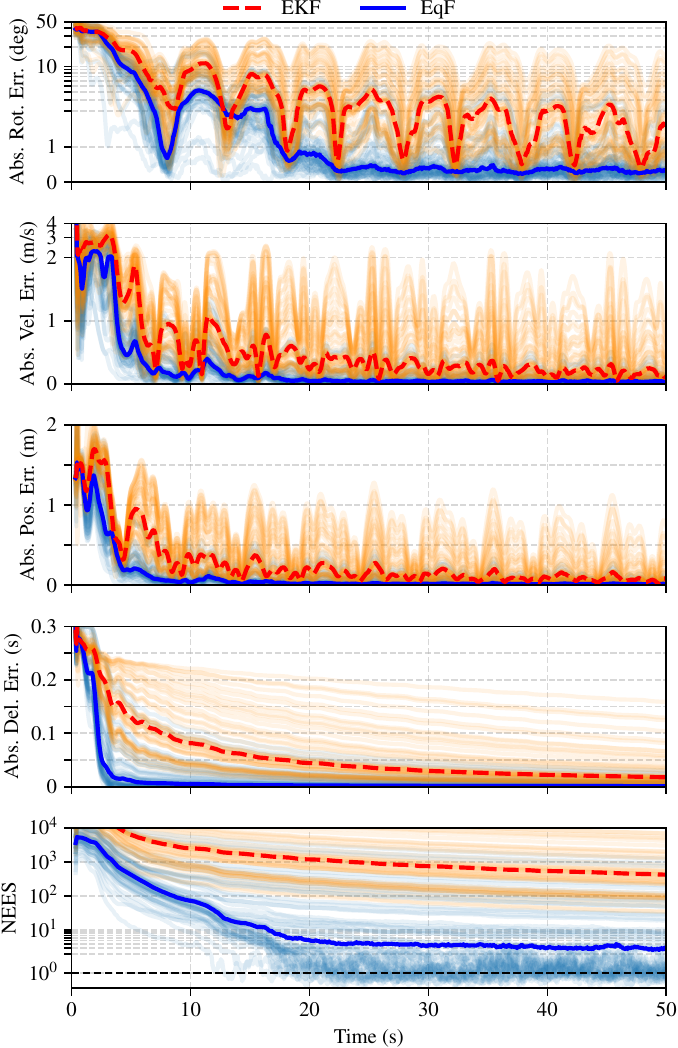}
        \end{minipage}%
    }
    \hfill
    \subfloat[$\delta = 500\,\mathrm{ms}$]{%
        \begin{minipage}[b]{0.245\textwidth}
            \includegraphics[width=\textwidth]{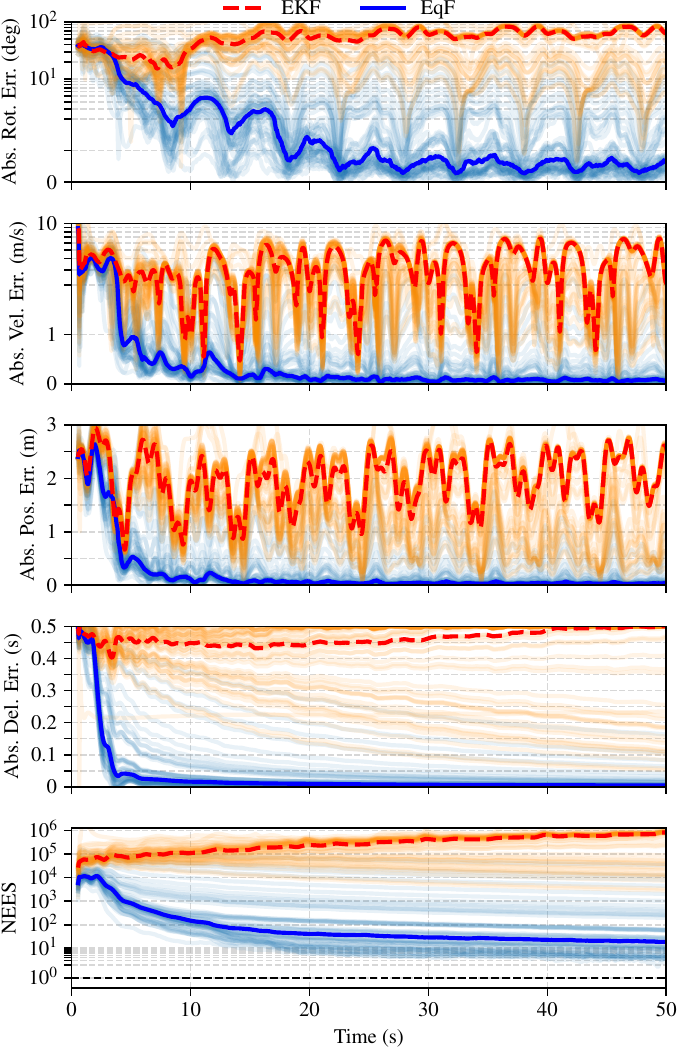}
        \end{minipage}%
    }
    \caption{Results of a Monte Carlo simulation with 50 different realizations of randomized initial states, process noise, and measurement noise, comparing EqF and EKF with \emph{online time-delay estimation} for increasing measurement delays: (a) $\delta = 100\,\mathrm{ms}$, (b) $\delta = 200\,\mathrm{ms}$, (c) $\delta = 300\,\mathrm{ms}$, and (d) $\delta = 500\,\mathrm{ms}$. 
    From top to bottom, the rows show the absolute rotation, velocity, position, and time-delay errors, followed by the NEES. Semi-transparent lines correspond to individual runs, while thick lines indicate the median across all runs. As the delay increases, the EKF exhibits growing oscillations, loss of consistency, and increased estimation error, ultimately diverging in the (d) case. In contrast, the EqF maintains stable, low-error, and consistent performance across all delays.}
    \label{fig:4_mc}
\end{figure*}

To complement the real-world experiments, we run simulations to provide a statistical comparison of the proposed EqF against the state-of-the-art EKF with online delay estimation, i.e., EKF (online).
We ran Monte Carlo simulations using an analytically generated trajectory consisting of circular motion with superimposed horizontal and vertical waves and varying attitude to excite all rotational axes.
Four delay scenarios were considered: 100\,ms, 200\,ms, 300\,ms, and 500\,ms.
Synthetic IMU measurements were generated at 200\,Hz, and GNSS position measurements at 20\,Hz, with a fixed time delay introduced according to the scenario.
For each scenario, 50 Monte Carlo realizations were performed by randomizing the initial states, process noise, and measurement noise.
The simulation results, reported in~\figref{4_mc}, highlight a clear distinction between the EqF and the EKF when the time delay is estimated online. 
Even for small measurement delays, the EKF exhibits inconsistencies, reflected in elevated NEES values, whereas the EqF maintains consistent performance across all delays.  
This behavior of the EKF aligns with the findings of~\citet{Kelly2021ASystems} showing that incorporating online time-delay estimation in classical robotic state-estimation filters introduces fundamental structural issues, leading to bias and inconsistency.  
In fact, as the delay increases, the EKF exhibits growing oscillations, loss of consistency, and increased estimation errors, ultimately diverging in the 500\,ms delay scenario.  
In contrast, the EqF built on the proposed geometric framework maintains accuracy and consistency, exhibiting full convergence in all tests.

\section{Conclusion}
\label{sec:conclusion}
This paper presented a geometric approach for modeling inertial motion with temporal delays based on Galilean symmetry.
The resulting mathematical framework enables spatial-temporal state estimation and offers a principled representation of INS with measurement delays.
Using this formulation, we derived an EqF that simultaneously estimates navigation states and unknown GNSS time lags.
We validated the approach with real-world flights of two UAVs experiencing GNSS delays of 90\,ms and 120\,ms.
The EqF accurately estimated both navigation states and time delays online, achieving RMSE performance comparable to an EKF with offline-estimated delay compensation, but without requiring tedious offline calibration.
Simulations provided further statistical comparison of our EqF against a state-of-the-art EKF with online time-delay estimation.
We investigated delays up to 500\,ms and observed that the EqF maintains accurate and consistent estimates, while the EKF remains inconsistent and exhibits growing errors with increasing delays, confirming previous findings~\cite{Kelly2021ASystems}.
These results demonstrate that exploiting Galilean symmetry within an EqF formulation provides a viable approach for joint navigation and delay estimation in GNSS-aided INS.
\section*{Acknowledgments}
The authors thank the ArduPilot community and in particular Andrew Tridgell for conducting the experimental flights.
They also thank Alessandro Fornasier for valuable discussions.
This work was supported by BMK under the grant agreement 913968 (SAMURAI), by the ARO W911NF-21-2-0245, and by the Franco-Australian International Research Project “Advancing Autonomy for Unmanned Robotic Systems” (IRP ARS).
The views and conclusions contained in this document are those of the authors only.
The U.S. Government is authorized to reproduce and distribute reprints for Government purposes.

\bibliographystyle{plainnat}
\bibliography{bibliography/references}

@article{Kelly2014ASensors,
    title = {\href{https://starslab.ca/wp-content/papercite-data/pdf/2014_kelly_general.pdf}{A General Framework for Temporal Calibration of Multiple Proprioceptive and Exteroceptive Sensors}},
    year = {2014},
    journal = {Springer Tracts in Advanced Robotics},
    author = {Kelly, Jonathan and Sukhatme, Gaurav S.},
    pages = {195--209},
    volume = {79},
    publisher = {Springer, Berlin, Heidelberg},
    isbn = {978-3-642-28572-1},
    doi = {10.1007/978-3-642-28572-1_14},
    issn = {1610-742X}
}

@article{Kelly2021ASystems,
    title = {\href{https://doi.org/10.1109/MFI52462.2021.9591176}{A Question of Time: Revisiting the Use of Recursive Filtering for Temporal Calibration of Multisensor Systems}},
    year = {2021},
    journal = {IEEE International Conference on Multisensor Fusion and Integration for Intelligent Systems},
    author = {Kelly, Jonathan and Grebe, Christopher and Giamou, Matthew},
    month = {11},
    publisher = {Institute of Electrical and Electronics Engineers Inc.},
    doi = {10.1109/MFI52462.2021.9591176},
    arxivId = {2106.00391v3}
}

@article{Frei2023ARendezvous,
    title = {\href{https://doi.org/10.1016/j.asr.2022.10.025}{A robust navigation filter fusing delayed measurements from multiple sensors and its application to spacecraft rendezvous}},
    year = {2023},
    journal = {Advances in Space Research},
    author = {Frei, Heike and Burri, Matthias and Rems, Florian and Risse, Eicke Alexander},
    number = {7},
    month = {10},
    pages = {2874--2900},
    volume = {72},
    publisher = {Pergamon},
    doi = {10.1016/J.ASR.2022.10.025},
    issn = {0273-1177}
}

@article{vanGoor2024ConstructiveMeasurements,
    title = {\href{https://doi.org/10.1016/j.ejcon.2024.101047}{Constructive synchronous observer design for inertial navigation with delayed GNSS measurements}},
    year = {2024},
    journal = {European Journal of Control},
    author = {van Goor, Pieter and Wickramasinghe, Punjaya and Hampsey, Matthew and Mahony, Robert},
    month = {11},
    pages = {101047},
    volume = {80},
    publisher = {Elsevier},
    doi = {10.1016/J.EJCON.2024.101047},
    issn = {0947-3580}
}

@article{Kelly2014DeterminingMeasurements,
    title = {\href{https://doi.org/10.1109/TRO.2014.2343073}{Determining the time delay between inertial and visual sensor measurements}},
    year = {2014},
    journal = {IEEE Transactions on Robotics},
    author = {Kelly, Jonathan and Roy, Nicholas and Sukhatme, Gaurav S.},
    number = {6},
    month = {12},
    pages = {1514--1523},
    volume = {30},
    publisher = {Institute of Electrical and Electronics Engineers Inc.},
    doi = {10.1109/TRO.2014.2343073},
    issn = {15523098}
}

@article{Kim2025EKF-BasedCalibration,
    title = {\href{https://doi.org/10.1109/LRA.2025.3575320}{EKF-Based Radar-Inertial Odometry With Online Temporal Calibration}},
    year = {2025},
    journal = {IEEE Robotics and Automation Letters},
    author = {Kim, Changseung and Bae, Geunsik and Shin, Woojae and Wang, Sen and Oh, Hyondong},
    number = {7},
    pages = {7230--7237},
    volume = {10},
    publisher = {Institute of Electrical and Electronics Engineers Inc.},
    doi = {10.1109/LRA.2025.3575320},
    issn = {23773766}
}

@article{Guo2023EnhancedIntegration,
    title = {\href{https://doi.org/10.1109/JSEN.2022.3223974}{Enhanced EKF-Based Time Calibration for GNSS/UWB Tight Integration}},
    year = {2023},
    journal = {IEEE Sensors Journal},
    author = {Guo, Yihan and Vouch, Oliviero and Zocca, Simone and Minetto, Alex and Dovis, Fabio},
    number = {1},
    month = {1},
    pages = {552--566},
    volume = {23},
    publisher = {Institute of Electrical and Electronics Engineers Inc.},
    doi = {10.1109/JSEN.2022.3223974},
    issn = {15581748}
}

@article{Mahony2025GalileanRobotics,
    title = {\href{https://arxiv.org/abs/2510.10468}{Galilean Symmetry in Robotics}},
    year = {2025},
    journal = {arXiv:2510.10468},
    author = {Mahony, Robert and Kelly, Jonathan and Weiss, Stephan},
    month = {10},
    arxivId = {2510.10468}
}

@article{Larsen1998IncorporationFilter,
    title = {\href{https://doi.org/10.1109/CDC.1998.761918}{Incorporation of time delayed measurements in a discrete-time Kalman filter}},
    year = {1998},
    journal = {Proceedings of the IEEE Conference on Decision and Control},
    author = {Larsen, Thomas Dall and Andersen, Nils A. and Ravn, Ole and Poulsen, Niels Kjolstad},
    pages = {3972--3977},
    volume = {4},
    publisher = {IEEE},
    doi = {10.1109/CDC.1998.761918},
    issn = {01912216}
}

@article{Nilsson2010JointSystems,
    title = {\href{https://doi.org/10.1109/ISSPA.2010.5605534}{Joint state and measurement time-delay estimation of nonlinear state space systems}},
    year = {2010},
    journal = {10th International Conference on Information Sciences, Signal Processing and their Applications, ISSPA 2010},
    author = {Nilsson, John Olof and Skog, Isaac and Handel, Peter},
    pages = {324--328},
    isbn = {9781424471676},
    doi = {10.1109/ISSPA.2010.5605534}
}

@article{Fatehi2017KalmanDelay,
    title = {\href{https://doi.org/10.1016/j.jprocont.2017.02.010}{Kalman filtering approach to multi-rate information fusion in the presence of irregular sampling rate and variable measurement delay}},
    year = {2017},
    journal = {Journal of Process Control},
    author = {Fatehi, Alireza and Huang, Biao},
    month = {5},
    pages = {15--25},
    volume = {53},
    publisher = {Elsevier},
    doi = {10.1016/J.JPROCONT.2017.02.010},
    issn = {0959-1524}
}

@article{Mina2025RemarksFiltering,
    title = {\href{https://arxiv.org/abs/2508.21260}{Remarks on stochastic cloning and delayed-state filtering}},
    year = {2025},
    journal = {arXiv:2508.21260},
    author = {Mina, Tara and Marinello, Lindsey and Christian, John},
    month = {8},
    isbn = {2508.21260v1},
    arxivId = {2508.21260}
}

@article{Khosravian2016StateMeasurements,
    title = {\href{https://doi.org/10.1016/j.automatica.2016.01.024}{State estimation for invariant systems on Lie groups with delayed output measurements}},
    year = {2016},
    journal = {Automatica},
    author = {Khosravian, Alireza and Trumpf, Jochen and Mahony, Robert and Hamel, Tarek},
    month = {6},
    pages = {254--265},
    volume = {68},
    publisher = {Pergamon},
    doi = {10.1016/J.AUTOMATICA.2016.01.024},
    issn = {0005-1098}
}

@article{Delama2025EquivariantApproach,
    title = {\href{https://doi.org/10.1109/LRA.2024.3511424}{Equivariant IMU Preintegration With Biases: A Galilean Group Approach}},
    year = {2025},
    journal = {IEEE Robotics and Automation Letters},
    author = {Delama, Giulio and Fornasier, Alessandro and Mahony, Robert and Weiss, Stephan},
    number = {1},
    pages = {724--731},
    volume = {10},
    publisher = {Institute of Electrical and Electronics Engineers Inc.},
    doi = {10.1109/LRA.2024.3511424},
    issn = {23773766},
    arxivId = {2411.05548}
}

@article{Kelly2023AllSGal3,
    title = {\href{https://arxiv.org/abs/2312.07555}{All About the Galilean Group SGal(3)}},
    year = {2023},
    journal = {arXiv:2312.07555},
    author = {Kelly, Jonathan},
    month = {12},
    arxivId = {2312.07555}
}

@article{vanGoor2022EquivariantEqF,
    title = {\href{https://doi.org/10.1109/TAC.2022.3194094}{Equivariant Filter (EqF)}},
    year = {2022},
    journal = {IEEE Transactions on Automatic Control},
    author = {van Goor, Pieter and Hamel, Tarek and Mahony, Robert},
    number = {6},
    month = {6},
    pages = {3501--3512},
    volume = {68},
    publisher = {Institute of Electrical and Electronics Engineers Inc.},
    doi = {10.1109/TAC.2022.3194094},
    issn = {15582523},
    arxivId = {2010.14666}
}

@article{Ge2022EquivariantSystems,
    title = {\href{https://doi.org/10.1109/CDC51059.2022.9992342}{Equivariant Filter Design for Discrete-time Systems}},
    year = {2022},
    journal = {Proceedings of the IEEE Conference on Decision and Control},
    author = {Ge, Yixiao and Van Goor, Pieter and Mahony, Robert},
    pages = {1243--1250},
    volume = {2022-December},
    publisher = {Institute of Electrical and Electronics Engineers Inc.},
    isbn = {9781665467612},
    doi = {10.1109/CDC51059.2022.9992342},
    issn = {25762370}
}

@article{Fornasier2022EquivariantBiases,
    title = {\href{https://doi.org/10.1109/ICRA46639.2022.9811778}{Equivariant Filter Design for Inertial Navigation Systems with Input Measurement Biases}},
    year = {2022},
    journal = {2022 International Conference on Robotics and Automation (ICRA)},
    author = {Fornasier, Alessandro and Ng, Yonhon and Mahony, Robert and Weiss, Stephan},
    month = {5},
    pages = {4333--4339},
    publisher = {IEEE},
    isbn = {978-1-7281-9681-7},
    doi = {10.1109/ICRA46639.2022.9811778}
}

@article{Fornasier2022OvercomingCalibration,
    title = {\href{https://doi.org/10.1109/LRA.2022.3210867}{Overcoming Bias: Equivariant Filter Design for Biased Attitude Estimation With Online Calibration}},
    year = {2022},
    journal = {IEEE Robotics and Automation Letters},
    author = {Fornasier, Alessandro and Ng, Yonhon and Brommer, Christian and Bohm, Christoph and Mahony, Robert and Weiss, Stephan},
    number = {4},
    month = {10},
    pages = {12118--12125},
    volume = {7},
    publisher = {Institute of Electrical and Electronics Engineers Inc.},
    doi = {10.1109/LRA.2022.3210867},
    issn = {23773766},
    arxivId = {2209.12038}
}

@article{Fornasier2023MSCEqF:Navigation,
    title = {\href{https://doi.org/10.1109/LRA.2023.3335775}{MSCEqF: A Multi State Constraint Equivariant Filter for Vision-aided Inertial Navigation}},
    year = {2023},
    journal = {IEEE Robotics and Automation Letters},
    author = {Fornasier, Alessandro and Goor, Pieter van and Allak, Eren and Mahony, Robert and Weiss, Stephan},
    month = {1},
    publisher = {Institute of Electrical and Electronics Engineers Inc.},
    doi = {10.1109/LRA.2023.3335775},
    issn = {23773766}
}

@article{Fornasier2024AnSystem,
    title = {\href{https://doi.org/10.1109/ICRA57147.2024.10611108}{An Equivariant Approach to Robust State Estimation for the ArduPilot Autopilot System}},
    year = {2024},
    journal = {Proceedings - IEEE International Conference on Robotics and Automation},
    author = {Fornasier, Alessandro and Ge, Yixiao and Van Goor, Pieter and Scheiber, Martin and Tridgell, Andrew and Mahony, Robert and Weiss, Stephan},
    pages = {11956--11962},
    publisher = {Institute of Electrical and Electronics Engineers Inc.},
    isbn = {9798350384574},
    doi = {10.1109/ICRA57147.2024.10611108},
    issn = {10504729}
}

@article{Fornasier2025EquivariantSystems,
    title = {\href{https://doi.org/10.1016/j.automatica.2025.112495}{Equivariant symmetries for inertial navigation systems}},
    year = {2025},
    journal = {Automatica},
    author = {Fornasier, Alessandro and Ge, Yixiao and van Goor, Pieter and Mahony, Robert and Weiss, Stephan},
    pages = {112495},
    volume = {181},
    doi = {10.1016/J.AUTOMATICA.2025.112495},
    issn = {0005-1098}
}

@inproceedings{Scheiber2023RevisitingApproach,
    title = {\href{https://doi.org/10.1109/ICAR58858.2023.10406552}{Revisiting Multi-GNSS Navigation for UAVs – An Equivariant Filtering Approach}},
    year = {2023},
    booktitle = {2023 21st International Conference on Advanced Robotics (ICAR)},
    author = {Scheiber, Martin and Fornasier, Alessandro and Brommer, Christian and Weiss, Stephan},
    month = {12},
    pages = {134--141},
    publisher = {IEEE},
    isbn = {979-8-3503-4229-1},
    doi = {10.1109/ICAR58858.2023.10406552}
}

@article{Mingyang2014OnlineAlgorithms,
    author = {Mingyang Li and Anastasios I. Mourikis},
    title = {\href{https://doi.org/10.1177/0278364913515286}{Online Temporal Calibration for Camera–IMU Systems: Theory and Algorithms}},
    journal = {The International Journal of Robotics Research},
    volume = {33},
    number = {7},
    pages = {947-964},
    year = {2014},
    doi = {10.1177/0278364913515286}
}

@article{Song2025unleashingpowerdiscretetimestate,
      title={\href{https://arxiv.org/abs/2509.12846}{Unleashing the Power of Discrete-Time State Representation: Ultrafast Target-based IMU-Camera Spatial-Temporal Calibration}}, 
      author={Junlin Song and Antoine Richard and Miguel Olivares-Mendez},
      year={2025},
      eprint={2509.12846},
      archivePrefix={arXiv},
      primaryClass={cs.RO},
      journal = {arXiv:2509.12846},
}

@article{GE2026106656,
title = {\href{https://www.sciencedirect.com/science/article/pii/S0967066125004186}{The difference between the left and right invariant extended Kalman filter}},
journal = {Control Engineering Practice},
volume = {167},
pages = {106656},
year = {2026},
issn = {0967-0661},
doi = {https://doi.org/10.1016/j.conengprac.2025.106656},
author = {Yixiao Ge and Giulio Delama and Martin Scheiber and Alessandro Fornasier and Pieter {van Goor} and Stephan Weiss and Robert Mahony}
}

\end{document}